%% file: main.tex
\documentclass[10pt,twocolumn,letterpaper]{article}

\usepackage{iccv}              
\usepackage{multirow} 
\usepackage{xcolor} 
\usepackage{makecell}
\usepackage[accsupp]{axessibility}

\input{preamble}

\usepackage{bm}

\definecolor{iccvblue}{rgb}{0.21,0.49,0.74}
\usepackage[pagebackref,breaklinks,colorlinks,allcolors=iccvblue]{hyperref}

\def\paperID{4741} 
\def\confName{ICCV}
\def\confYear{2025}

\title{InvRGB+L: Inverse Rendering of Complex Scenes with  \\ Unified Color and LiDAR Reflectance Modeling}

\author{
Xiaoxue Chen$^{1,2}$ \quad
Bhargav Chandaka$^{2}$ \quad
Chih-Hao Lin$^{2}$ \quad
Ya-Qin Zhang$^{1}$ \quad
David Forsyth$^{2}$ \quad \\
Hao Zhao$^{1,3}$ \quad
Shenlong Wang$^{2}$ \\
$^1$AIR, Tsinghua University \quad
$^2$University of Illinois Urbana-Champaign \quad
$^3$BAAI \\
}

\begin{document}

\twocolumn[{%
\maketitle
\vspace{-1cm}
\renewcommand\twocolumn[1][]{#1}%
\begin{center}
    \centering
    \includegraphics[width=1\linewidth]{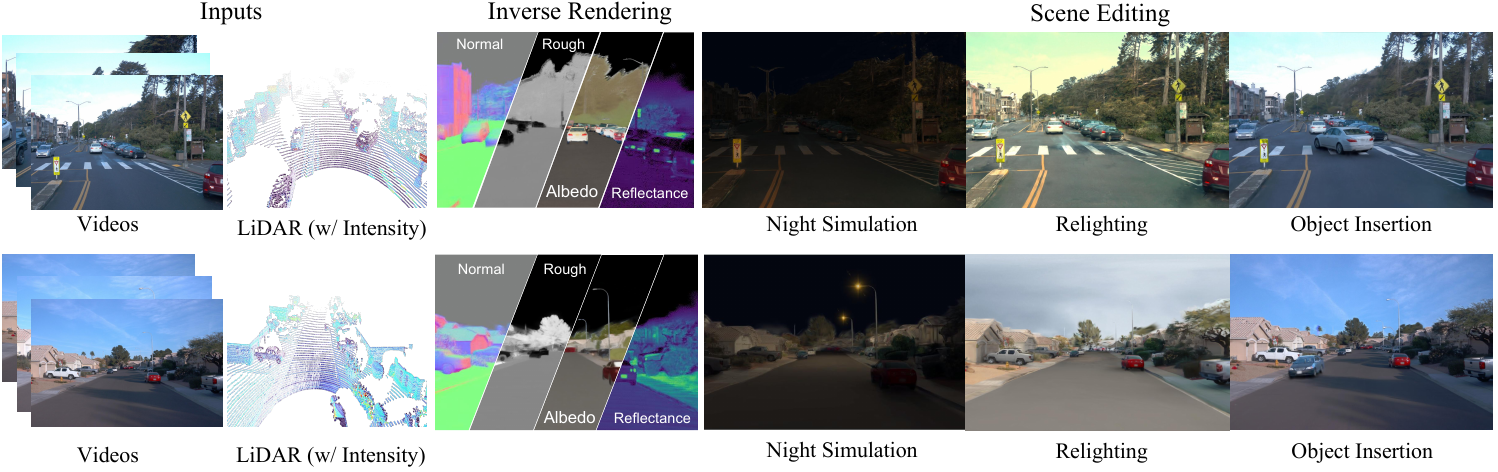}
    \vspace{-5mm}
    \captionof{figure}{{\bf Overview:} \ours~takes RGB and LiDAR sequences as input and outputs a 3D scene with high-fidelity geometry, consistent albedo across RGB and LiDAR spectra, and roughness. Our representation enables photorealistic object insertion and night simulations. 
    }
    \label{fig:teaser}
\end{center}
}]

\input{sec/0_abstract}    
\input{sec/1_intro-daf-1}

\input{sec/2_relatedworks-daf-1}
\input{sec/3_lidar-daf-1}

\input{sec/4_inverse_rendering_framework-daf-2}

\input{sec/5_experiments-daf-1}

{
    \small

\input{main.bbl}
}
\input{sec/6_supplementary}


\end{document}

%% file: preamble.tex
\newcommand{\ours}{{InvRGB+L}}

\definecolor{darkorange}{rgb}{1,0.55,0}
\definecolor{darkgreen}{rgb}{0,0.5,0}
\definecolor{darkgray}{rgb}{0.5,0.5,0.5}

%% file: sec/0_abstract.tex
\begin{abstract}
We present \ours, a novel inverse rendering model that reconstructs large, relightable, and dynamic scenes from a single RGB+LiDAR sequence. 
Conventional inverse graphics methods rely primarily on RGB observations and use LiDAR mainly for geometric information, often resulting in suboptimal material estimates due to visible light interference. 
We find that LiDAR’s intensity values—captured with active illumination in a different spectral range—offer complementary cues for robust material estimation under variable lighting. 
Inspired by this, \ours~leverages LiDAR intensity cues to overcome challenges inherent in RGB-centric inverse graphics through two key innovations: (1) a novel physics-based LiDAR shading model and (2) RGB–LiDAR material consistency losses. 
The model produces novel-view RGB and LiDAR renderings of urban and indoor scenes and supports relighting, night simulations, and dynamic object insertions—achieving results that surpass current state-of-the-art methods in both scene-level urban inverse rendering and LiDAR simulation.

\end{abstract}

%% file: sec/1_intro-daf-1.tex
\section{Introduction}
\label{sec:intro}

\begin{figure}[t]
\begin{center}
\includegraphics[width=0.99\linewidth]{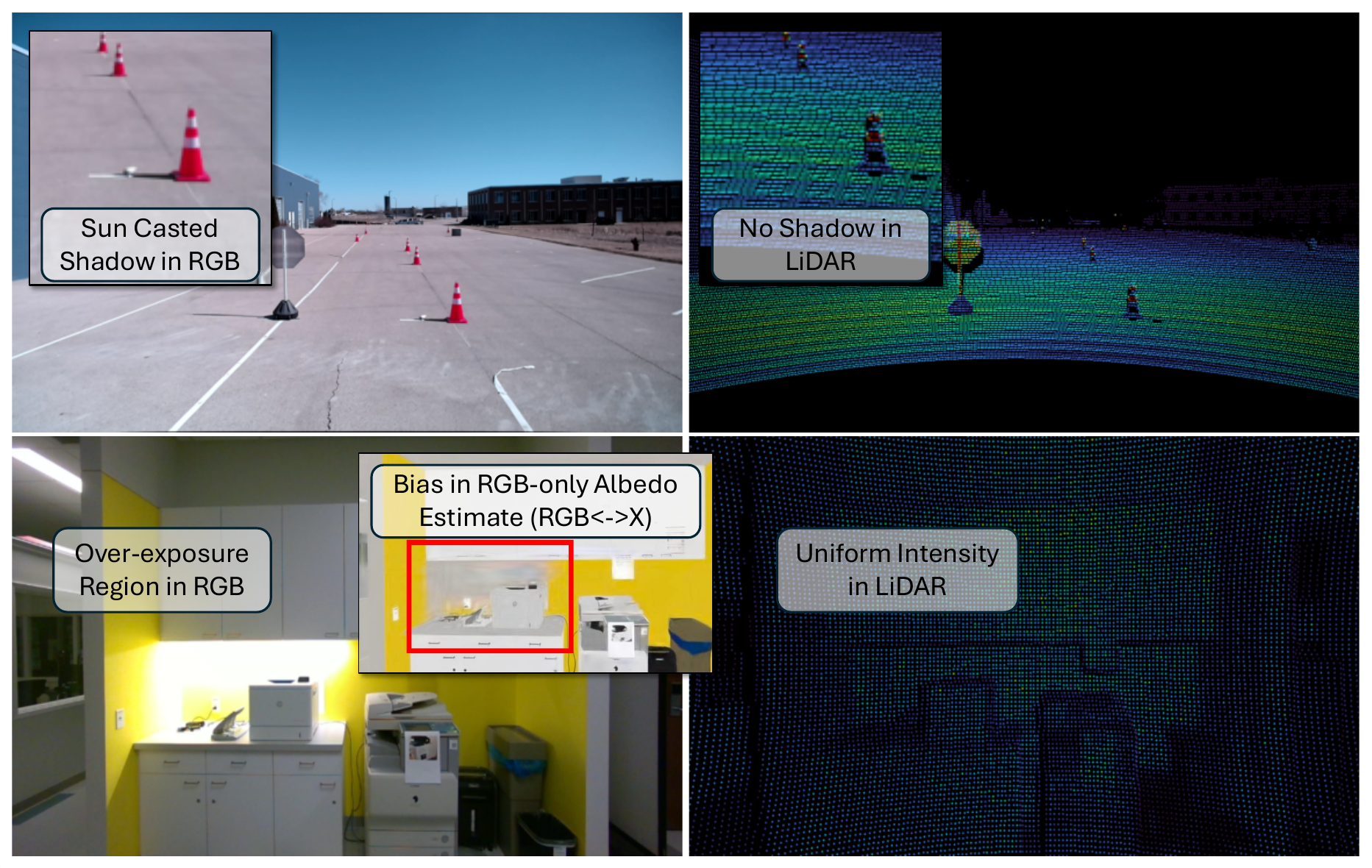}
\end{center}
\vspace{-5mm}
\caption{{\bf Key insight:} LiDAR reflectance is less affected by environmental lighting than color images, making it an excellent complement for inverse graphics. Top: Cast shadows in color images do not appear in LiDAR reflectance; Bottom: an overexposed yellow wall shows uniform reflectance in the LiDAR spectrum. }
\label{fig:insight}
\vspace{-1em}
\end{figure}

{\bf Inverse rendering} is challenging because image observations are wildly ambiguous.
The same image can be interpreted as a yellow wall lit by white light or as a wall that is half yellow and half white (Fig.~\ref{fig:insight});
a dark region might be interpreted as a wet area or as a cast shadow. Errors like these
in material recovery result in scene renderings that can be jarringly bad.  Even strong material priors only partially
mitigate these ambiguities (Fig.~\ref{fig:insight}, middle).    

{\it LiDAR intensity provides strong cues for inverse rendering}. LiDAR sensors emit laser pulses that
reflect off surfaces.  As is well known, time-of-flight yields geometry.  We demonstrate the returned power
(analogous to RGB intensity, Fig.\ref{fig:insight}) provides rich surface material information. LiDAR derived material
cues are extremely robust to wide changes in illumination conditions, because there is very little cross-talk between
the narrow-band infrared used by LiDAR and typical illuminants indoors and outdoors.   But material properties change
very slowly with wavelength.  So LiDAR returned power can, for example,  tell that the albedo of the wall in
Fig~\ref{fig:insight} is the same in the darker and lighter regions.  We show that LiDAR intensity is a powerful cue
that disentangles material properties and illumination effects in ways that complement SOTA methods for RGB  data.


{\bf \ours} is a novel inverse rendering framework that reconstructs large, relightable, and dynamic scenes from a
single RGB+LiDAR sequence. {\bf \ours} infers geometry, illumination, and materials using
{\it LiDAR intensity observations with color images together}, exploiting two novel technical
contributions: (1) a physics-based LiDAR reflectance model that—unlike conventional reflectance models—explicitly
accounts for surface specularity, and (2) a joint RGB–LiDAR material consistency loss that models the relationship
between visible and LiDAR's infrared observations.

Experiments show that our model produces novel-view RGB and LiDAR
renderings for both urban and indoor scenes accurately while also supporting realistic relighting, night simulations,
and dynamic object insertions.  Our method surpasses current state-of-the-art approaches in scene-level urban inverse
rendering and novel-view LiDAR simulation in qualitative and quantitative comparisons.

%% file: sec/2_relatedworks-daf-1.tex
\section{Related Works}


\textbf{Inverse Rendering} recovers scene properties like geometry \cite{huang2023neural_1, wang2024dust3r},
materials \cite{dorney2001material, meka2018lime}, and lighting \cite{zhan2021emlight, wang2022stylelight,
  phongthawee2024diffusionlight} from sensor data.  Light-surface interactions make the problem wildly ambiguous.
Data-driven methods use dense prediction networks \cite{yu2019inverserendernet, sengupta2019neural, chen2023dpf,
  zhu2022irisformer,zhu2022learning} and diffusion models \cite{fu2024geowizard, kocsis2024intrinsic, Zeng_2024,
  liang2025diffusionrenderer} to predict intrinsic properties.  The absence of an explicit physical model
can result in unrealistic outcomes.  Physics-based methods leverage 3D representations like NeRF
\cite{wang2023neural, zhu2023i2, jin2023tensoir, chen2023nerrf, lin2023urbanir, chen2024time,wu2023factorized} or 3D-GS \cite{shi2023gir,chen2024rgm,
  gao2024relightable, liang2024gs} to model geometry, then use differentiable PBR rendering to infer materials and
lighting.  Ambiguities remain, so priors are needed to constrain the solution space.

There exist methods that incorporate LiDAR cues~\cite{wang2023neural, pun2023neural}, but these
do not exploit LiDAR intensity.  All methods struggle with dynamic environments.
In contrast, we use LiDAR intensity as a powerful cue to material properties
and our method operates in dynamic environments.


\textbf{LiDAR Simulation} generates synthetic LiDAR data from existing observations to create new views or
counterfactual scenarios. Geometry simulation has
been tackled using point clouds \cite{li2023pcgen}, surfels \cite{manivasagam2020lidarsim}, NeRF
\cite{tonderski2024neurad, huang2023neural, tao2024lidar, wu2024dynamic, tao2024alignmif,yan2024street,yang2023unisim}, and 3DGS \cite{chen2024lidar, chen2024omnire}
as scene representations. Intensity simulation methods rely on lookup tables \cite{cheng2022generalized, manivasagam2020lidarsim} or encoded intensity fields
\cite{huang2023neural, chen2024lidar}. Many approaches mimic LiDAR ray-drop characteristics, but neglect
the physics of LiDAR reflectance.  In contrast, we show powerful inferences can be rooted in this physics; further, we show
close attention to LiDAR physics produces better simulations.  Work that models LiDAR reflection empirically
\cite{vacek2021learning, wu2021airborne, viswanath2023off, viswanath2024reflectivity} 
assumes Lambertian surfaces. In contrast, we offer a novel formulation incorporating a
specular term. Current methods produce sparse maps.  In contrast we show that
joint LiDAR-RGB inference results in dense, accurate maps.

%% file: sec/3_lidar-daf-1.tex
\begin{figure*}[t]
  \centering
  \includegraphics[width=1\textwidth]{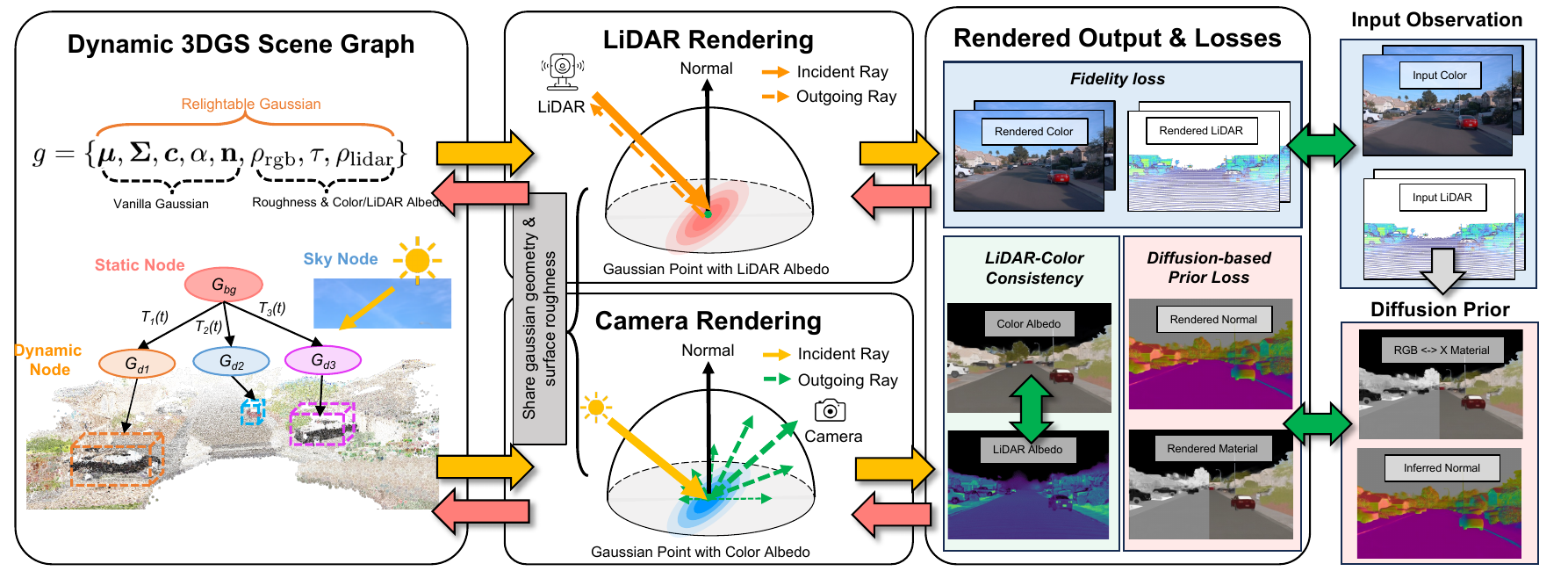}
  \caption{
  \textbf{Overall architecture}. We represent the scene as a dynamic, relightable 3DGS scene graph, consisting of a static node for the background, a set of dynamic nodes for movable objects, and a sky node to model illumination.
Our scene can generate realistic LiDAR and camera footage via physically based forward rendering modules.
Scene parameters are inferred through an inverse rendering process using backpropagation, minimizing discrepancies between rendered results and observations (as well as priors) while maximizing self-consistency.
\textbf{\color{darkorange} orange arrow}: forward rendering process;
\textbf{\color{darkgray} gray arrow}: diffusion-based normal and material prior inference;
\textbf{\color{red} red arrow}: backpropagation for inverse rendering;
\textbf{\color{darkgreen} green arrow}: loss computation.
  }
  \label{fig:main}
\end{figure*}

\section{Physics-based LiDAR Reflectance Model}

LiDAR  follows  the rendering equation~\cite{kajiya1986rendering} and we assume no in or out scattering,
so the reflected radiance is:
\begin{align}
  \label{eq:rendering}
L_r(\mathbf{x},\boldsymbol{\omega}_o) = \int_{\Omega} f_r(\mathbf{x}, \boldsymbol{\omega}_i, \boldsymbol{\omega}_o) L_i(\mathbf{x}, \boldsymbol{\omega}_i) (\mathbf{n} \cdot \boldsymbol{\omega}_i) d\boldsymbol{\omega}_i,
\end{align}  
where $ \mathbf{x} $ is the surface point, $ \mathbf{n} $ is the surface normal, $ \boldsymbol{\omega}_i $ and $ \boldsymbol{\omega}_o $ are incident and outgoing ray directions, $ L_i $ is the incident radiance, and $ f_r $ is the BRDF at $ \mathbf{x}$. 

LiDAR pulses are narrow and directional, so
$L_i( \mathbf{x}, \boldsymbol{\omega}_i)$ can be modelled as a constant value
in a very narrow beam around  $\boldsymbol{\omega}_0$ (Fig.~\ref{fig:main} middle).  Energy disperses, so
the radiance at $\mathbf{x}$ will be $L_i( \mathbf{x}, \boldsymbol{\omega}_i) \propto \frac{P_e}{d^2}$, where $ P_e $ is the emitted power
and $ d $ is the distance to $ \mathbf{x} $. The returned beam is narrow and the sensor responds to radiance, so
the sensor response is given by
$I(x,\boldsymbol{\omega}_o) \propto L_r(\mathbf{x}, \boldsymbol{\omega}_o) \propto f_r(\mathbf{x},
\boldsymbol{\omega}_o, \boldsymbol{\omega}_o) \frac{P_e \cos\theta}{d^2}, \label{eq: lidar intensity}$   
where $ \theta $ is the angle between $ \boldsymbol{\omega}_i $ and $ \mathbf{n} $.  

Existing models~\cite{huang2023neural,viswanath2024reflectivity,vacek2021learning,manivasagam2020lidarsim} assume
Lambertian (diffuse) surfaces,
where $f_r(\mathbf{x}, \boldsymbol{\omega}_o, \boldsymbol{\omega}_o)$ is constant $\rho_{\mathrm{lidar}}/\pi$, making 
$ I \propto \frac{\rho_\mathrm{lidar} P_e \cos\theta}{d^2}$,
where $\rho_\mathrm{lidar}$ represents the surface reflectance ( LiDAR  albedo).
However, this model fails to explain many real-world phenomena, such as the
spotlight reflectance on metallic surfaces (e.g., cars) and water foundations. 

We extend the LiDAR reflectance model by incorporating the Cook-Torrance BRDF~\cite{cook1982reflectance}, so
$ f_r = f_d + f_s $, where $ f_d = \frac{\rho_{\rm  lidar}}{\pi} $ is the diffuse term, and $ f_s $ is the specular
term.  Surface roughness $ \tau $ and angle $ \theta $ interact, yielding
$
f_s \frac{F_0 \tau^2 \mathrm{min}(1, 2cos^2\theta)}{4 \pi \cos^2\theta \left( \cos^2\theta (\tau^2 - 1) + 1 \right)^2}  $,
with fresnel term $ F_0 = 0.04 $. This specular component is a special case of the microfacet model, assuming the same incident and
outgoing ray angles. Substitution yields:
\begin{align}
I \propto  \left(\rho_\mathrm{lidar} + \frac{F_0 \tau^2 \mathrm{min}(1, 2cos^2\theta)}{4 \cos^2\theta \left( \cos^2\theta (\tau^2 - 1) + 1 \right)^2}  \right) \frac{P_e \cos\theta}{\pi d^2}. 
\label{eq: lidar intensity}
\end{align}  
By explicitly modeling specularity, our LiDAR reflectance model aligns with commonly used RGB-based shading models, enabling a unified framework for joint LiDAR and RGB inverse rendering in the following section. Refer to the supplementary material for details.

%% file: sec/4_inverse_rendering_framework-daf-2.tex
\section{Method}
\label{sec:method}



We recover a \textbf{relightable 4D scene representation} that
encodes geometry, color, LiDAR reflectance, and an HDR illumination model from 
an input video sequence $\{ \mathbf{C}_t \in \mathbb{R}^{W\times H \times 3} \}_{t=0}^T$, LiDAR sequences $\{
\mathbf{P}_t \in \mathbb{R}^{N \times 3} \}_{t=0}^T$ with intensity maps $\{ \mathbf{I}_t \in \mathbb{R}^{W\times H
  \times 3} \}_{t=0}^T$, and their corresponding poses $\{\boldsymbol{\xi}_t \in \mathbb{SE}(3)\}_{t=0}^T$ captured
under a single illumination environment.   We represent the scene as a dynamic scene
graph where each node is a 3D Gaussian encoding geometry, opacity, and intrinsic material
properties for both LiDAR and camera modalities
(Sec.~\ref{sec:scene}).  Forward rendering produces RGB
imagery and LiDAR intensity maps from a camera pose,
a scene graph and a physical model (Sec.~\ref{sec:forward}).  Inference adjusts scene parameters
to produce renderings that are like observed data; our inference
procedure introduce a novel albedo-consistency loss that synergizes RGB and LiDAR cues for joint
reasoning (Sec.~\ref{sec:rgbl inverse rendering}).
The architecture is presented in Fig. \ref{fig:main}. 

\subsection{Relightable Scene Representation}
\label{sec:scene}


\paragraph{Dynamic Scene Graph}  We use a dynamic scene graph $\mathbf{S}$,
where movable objects and backgrounds are explicitly represented as graph nodes.   The representation is
built out of Gaussian primitives as in 3D-GS.  Each primitive \( g(\boldsymbol{x})\) is defined by
$ g(\boldsymbol{x}) = e^{-\frac{1}{2} (\boldsymbol{x}-\boldsymbol{\mu})^T {\boldsymbol{\Sigma}}^{-1}
  (\boldsymbol{x}-\boldsymbol{\mu})} $, where \(\ \boldsymbol{x} \in \mathbb{R}^{3}\) is a 3D coordinate, \(\rm
\boldsymbol{\mu} \in \mathbb{R}^{3}\) stands for the mean of the Gaussian, and \(\ \boldsymbol{\Sigma} \in \mathbb{R}^{3
  \times 3}\) is the covariance matrix. We initialize the 3D means with LiDAR points for accurate geometry.  
In contrast, previous work \cite{wang2023neural, jin2023tensoir, lin2023urbanir} focuses on static scenes.

The background (eg roads and buildings) is modelled with a set of static Gaussians \(\mathbf{G}_{bg}\).  Moving objects are represented with
dynamic nodes \(\{\mathbf{G}'_{1}, \mathbf{G}'_{2}, ..., \mathbf{G}'_{N}\}\), where \(N\) is the number of objects.
Each object is represented as a 3D-GS in its local coordinate system. To place them in the dynamic scene, we apply a
pose transformation \(\mathbf{T}_k(t) \in \mathbb{SE}(3)\), where \(k\) is the object index and t is the timestamp. The
transformed Gaussian set is then formulated as \( \mathbf{G}_{k}(t) = \mathbf{T}_k(t) \cdot \mathbf{G}'_{k}\).

\paragraph{Illumination Model} We model the sky node with spherical harmonic (SH) illumination to approximate the global
lighting from the sky dome. We parameterize lighting with  SH coefficients up to 3rd-order
$\rm \mathbf{L}_{sky} \in  \mathbb{R}^{16 \times 3}$, and define the sky lighting from incident direction \(\omega_i\) as
\(\rm \mathbf{L}_{sky} (\boldsymbol{\omega}_i)\).  Sky lighting fails to model
sharp shadows.  Additionally, we use a learnable sun light $\mathbf{L}_{sun}=\{\boldsymbol{\omega}_{sun},I_{sun}\}$ to explicitly
model directional sunlight, where $\boldsymbol{\omega}_{sun}$ is the sunlight direction and $I_{sun}$ is the sunlight intensity.  

\paragraph{Relightable Gaussian} Each of our 3D Gaussian primitives  is associated with intrinsic parameters,
enabling geometry and material estimation during 3D-GS optimization.  We adopt the Cook-Torrance BRDF for RGB image
rendering (as in the LiDAR reflectance model), so the BRDF parametrization is
\( f_r(\boldsymbol{\mu}, \boldsymbol{\omega}_i, \boldsymbol{\omega}_o;  \mathbf{n},  \rho_\mathrm{rgb}, \tau) \),
where parameters are: \( \mathbf{n} \in \mathbb{S}^2 \) (surface normal);
\( {\rho_\mathrm{rgb}} \in [0,1]^3 \) (diffuse albedo); and \( \tau \in [0,1] \)
(surface roughness).  Surface normal and surface roughness will be the same at visible and LiDAR wavelengths,
but diffuse albedo may not be.  We denote LiDAR albedo in the physical reflectance model as  \(
 \rho_\mathrm{lidar} \in [0,1] \).  For each Gaussian primitive \( g({x}) \), these parameters are associated to model the material
 properties, so: $g =  \{\boldsymbol{\mu},
 \boldsymbol{\Sigma},\boldsymbol{c},\alpha , \mathbf{n}, \rho_\mathrm{rgb}, \tau, \rho_\mathrm{lidar} \}$.

The entire scene representation at timestamp \(t\) is then \(\mathbf{S} = \{ \mathbf{G}_\mathrm{bg}, \mathbf{G}_{k}(t),
\mathbf{L}_\mathrm{sky}, \mathbf{L}_\mathrm{sun}\}\). 



\subsection{Physics-based Forward Rendering}\label{sec:forward}

Physics-based forward rendering of the scene serves as the foundation
for inverse modeling to estimate scene parameters and supports
downstream applications such as relighting and insertion rendering.

\paragraph{Camera Rendering} We adopt a BVH-based ray tracer \cite{gao2024relightable}, denoted as \(
\text{Tracer}(\cdot) \) to trace the visibility for each Gaussian. For an incident direction \( \boldsymbol{\omega}_i
\), the visibility \( \rm v(\boldsymbol{\omega}_i) = \text{Tracer}(\boldsymbol{\omega}_i; \mathbf{S}) \) indicates
whether the Gaussian receives direct illumination from the sky. If a ray from \( g \) toward \( \boldsymbol{\omega}_i \)
intersects another object before reaching the sky dome, \( \rm v(\boldsymbol{\omega}_i) = 0 \); otherwise, it is
directly lit, and \( \rm v(\boldsymbol{\omega}_i) = 1 \). However, in urban scenes, restricting ray tracing to only
visible objects can lead to incomplete shadowing, as occluded objects outside the field of view may also cast
shadows. To address this, we introduce an sun visibility parameter \( \rm v_{sun} \) for each $g$, which indicates
whether a Gaussian is directly lit by sunlight \( \mathbf{L}_{\text{sun}} \) from direction \(
\boldsymbol{\omega}_{\text{sun}} \). 

The PBR color for each Gaussian primitive can be computed using the rendering equation. We employ Monte Carlo sampling
to generate M incident ray directions. Consequently, the estimated PBR color $\rm \hat{c}$ of a Gaussian primitive $g$
for view direction $\omega_o$ is: $  \rm \hat{c}(\mathbf{\omega}_o)= \frac{1}{M} \sum_{i=1}^{M} [v_{sun} I_{sun}
  (\boldsymbol{\omega}_{sun}\cdot \boldsymbol{n}) + \rm v(\boldsymbol{\omega}_i) f_r(\mu, \boldsymbol{\omega}_i,
  \boldsymbol{\omega}_o;  \mathbf{n}, \rho_\mathrm{rgb}, \tau_\mathrm{rough}) \mathbf{L}_{sky}(\boldsymbol{\omega}_i)
  \cos\theta ] . $ The first term is the sunlight while the second term is the incident lighting from the sky dome. We
then render the scene graph $\mathbf{S}$  into the image space through $\rm \alpha$-blending as $\rm \mathbf{\hat{C}} =
\sum_{j} \alpha_j \hat{c}_j \prod_{k<j} (1 - \alpha_k)$. Additionally, we render all the attributes into corresponding
maps (e.g., normal map \(\rm  \mathbf{N} \), albedo map \( \rm \mathbf{B}_\mathrm{rgb} \), roughness map \( \rm
\mathbf{R} \) ) using $\rm \alpha$-blending. The camera rendering results of scene graph S are defined as:
$\mathtt{render}_\mathrm{rgb}(\mathbf{S}) = \rm \{\mathbf{\hat{C}},\mathbf{N},\mathbf{B}_\mathrm{rgb},\mathbf{R}\}$. 



\paragraph{LiDAR Rendering}  Given the reflectance parameter \( \rho_{\mathrm{lidar}_j} \) and the LiDAR reflectance
model in Eq.~\ref{eq: lidar intensity}, we compute the intensity value for each Gaussian. Since LiDAR sensing involves a
single incident ray—the laser itself—no sampling is required. The intensity \( I \) for a Gaussian is given by
Eq. \ref{eq: lidar intensity}, 
where \( d \) and \( \omega_o \) represent the distance and direction from the LiDAR origin to the Gaussian center \(
\mu \), and $\rm cos\theta = \mathbf{n} \cdot \mathbf{\omega_o}$.We assume that the laser energy of each LiDAR channel
is calibrated, setting \( P_e = 1 \).  Finally, we render both the intensity map \( \rm \hat{I} \) and the reflectance
map \( \rm \mathbf{B}_{lidar} \) into image space, defining the LiDAR rendering process as:
$\mathtt{render}_\mathrm{lidar}(\mathbf{S}) = \rm \{\mathbf{\hat{I}}, \mathbf{B}_{lidar}\}. $

\subsection{Inverse Rendering with RGB+L}
\label{sec:rgbl inverse rendering}


\paragraph{Problem Formulation} We must infer scene parameters -- geometry, material properties, illumination, and LiDAR
reflectance --  from both RGB and LiDAR data.  The overall loss function for optimizing the scene graph $\mathbf{S}$ is:
\begin{align}
    \min_{\mathbf{S}} \underbrace{\mathcal{L}_\mathrm{lidar} + \mathcal{L}_\mathrm{rgb}}_\mathrm{fidelity} + \underbrace{\mathcal{L}_\mathrm{nor} + \mathcal{L}_\mathrm{mat}}_\mathrm{diffusion\ prior} + \underbrace{\mathcal{L}_\mathrm{rgb\rightarrow lidar} + \mathcal{L}_\mathrm{lidar\rightarrow rgb}}_\mathrm{rgb-lidar\ consistency}
\end{align}
    {\bf Fidelity losses} for LiDAR and RGB are:
\begin{align} 
\rm \mathcal{L}_{rgb} = \Vert \hat{\mathbf{C}}-\mathbf{C}  \Vert_2^2,
 \quad \mathcal{L}_{lidar} = \Vert (\hat{\mathbf{I}}-\mathbf{I})\cdot \mathbf{M}_{lidar}  \Vert_2^2
\end{align}
where $\mathbf{C}$ is the ground-truth image and 
$ \mathbf{M}_{lidar}$ is a mask to account for sparseness in LiDAR intensity observations.  The mask is obtained by
thresholding.

\paragraph{Diffusion-based Prior} We mitigate the ambiguity in material inference by using monocular geometric and
material cues from pre-trained models.  We use Geowizard\cite{fu2024geowizard} and
RGB-X\cite{Zeng_2024} to preprocess multi-view training images, extracting pseudo normal and material labels, written
$\rm \mathbf{\hat{N}}$ and $\rm \mathbf{\hat{M}} = \{\mathbf{\hat{B}}_\mathrm{rgb},\mathbf{\hat{R}}\}$ respectively.
These guide inference through losses:
\begin{align}
\rm \mathcal{L}_{nor} = \Vert \mathbf{N}-\mathbf{\hat{N}}  \Vert_2^2,  \quad
\rm \mathcal{L}_{mat} = \Vert \mathbf{M}-\mathbf{\hat{M}}  \Vert_2^2.
\end{align}

\paragraph{RGB-LiDAR Albedo Consistency Loss} 



Spectral reflectance (which affects RGB images) and LiDAR albedo are strongly spatially correlated because each
is an epiphenomenon of material microstructure (see also~\cite{MalWand,Krinov}).  Surfaces with similar
spectral reflectance will tend to have similar LiDAR albedo and vice versa. We introduce two consistency
constraints that enforce the correlation between albedo and reflectance.

LiDAR intensity maps are inherently sparse and incomplete, but  spectral reflectance is a dense
signal. We enforce a neighborhood smoothness constraint that propagates the sparse LiDAR albedo
values into a dense map $\rm B_{lidar}$, under the assumption that it should exhibit similar smoothness as the reflectance
map $\mathbf{B}_\mathrm{rgb}$. Specifically, we adopt a bilateral smoothness loss: 
\begin{align} 
{\mathcal{L}_{\rm rgb\rightarrow lidar}} =   \sum_{q \in N(p)} |\mathbf{B}_\mathrm{lidar_p}-\mathbf{B}_\mathrm{lidar_q}| \cdot w(\mathbf{B}_\mathrm{rgb_p}, \mathbf{B}_\mathrm{rgb_q}), \end{align}
Here, $p$ and $q$ denote indexes of neighboring pixels,  and the weighting function is given by $ w(
\mathbf{B}_\mathrm{rgb_p}, \mathbf{B}_\mathrm{rgb_q}) = \exp \left( -\frac{ (\mathbf{B}_\mathrm{rgb_p} -
  \mathbf{B}_\mathrm{rgb_q})^2}{\sigma^2} \right)$. $\sigma$ is a hyperparameter controlling the sensitivity  to
spectral reflectance differences, ensuring smooth propagation of reflectance while preserving material boundaries. 

LiDAR albedo estimates are independent of external lighting conditions, so are a powerful cue for correcting
errors in reflectance estimates caused by lighting.  Assume surfaces with similar albedo will have similar spectral
reflectance; then we can impose a  regional consistency on spectral reflectance by: 
\begin{align} 
\rm \mathcal{L}_{lidar\rightarrow rgb} = \sum_\Omega \mathrm{var}(\mathbf{B}_{rgb_\Omega} | \mathbf{B}_{lidar_\Omega}) , \end{align}
where  $\Omega$ is a set of regions operating as superpixels (obtained using SAM \cite{kirillov2023segment}) within the LiDAR albedo map
$\rm \mathbf{B}_{lidar}$.  $\mathbf{B}_{rgb_\Omega}$ and $\mathbf{B}_{lidar_\Omega}$ denote the sets 
of spectral reflectance and albedo values within the same region $\Omega$.  




\paragraph{Optimization} We use a two-stage optimization process. In the first stage, we supervise the scene graph
without the consistency loss to obtain the initial Gaussians (only geometry, opacity and non-relightable colors) and
scene graph topology.  In the second stage, we fix the geometry and opacity of the Gaussians, and refine the intrinsic
material properties and lighting through joint optimization. For details please refer to the supplementary materials.

%% file: sec/5_experiments-daf-1.tex
\section{Experiment}

\begin{figure}[t]
  \centering
  \includegraphics[width=1\linewidth]{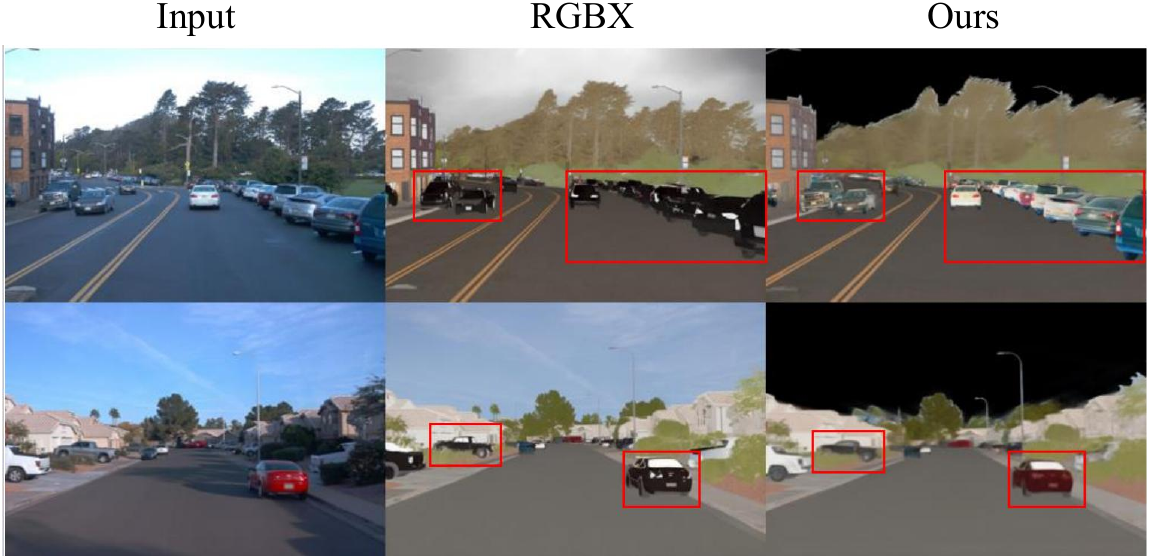}
  \caption{{\bf Our estimated spectral reflectance vs RGB$\leftrightarrow$X.}
Compared to the latest generative diffusion prior~\cite{Zeng_2024}, our estimated spectral reflectance better reflects the vehicle's paint color and is more robust to cast shadows.} 
  \label{fig:mono}
\end{figure}

\subsection{Experiment Protocols}

\paragraph{Datasets}
We conduct experiments on the Waymo Open Dataset \cite{sun2020scalability}, which provides RGB images from five cameras and 64-beam LiDAR data (including intensity). uring training, we use only one camera and its corresponding LiDAR sequence. Since each scene is recorded only once, the dataset does not support quantitative relighting evaluation. To address this, we collected an additional driving scene recorded at different times of the day, capturing varying illumination conditions. We also captured an indoor scene under an artificial lighting environment to verify the effectiveness of our albedo-reflectance consistency.

\paragraph{Evaluation Metrics}
For image comparisons, we use PSNR, SSIM~\cite{wang2004image}, and LPIPS~\cite{Zhang2018}. For LiDAR intensity simulation, we evaluate using RMSE.

\begin{figure}[t]
  \centering
  \includegraphics[width=1\linewidth]{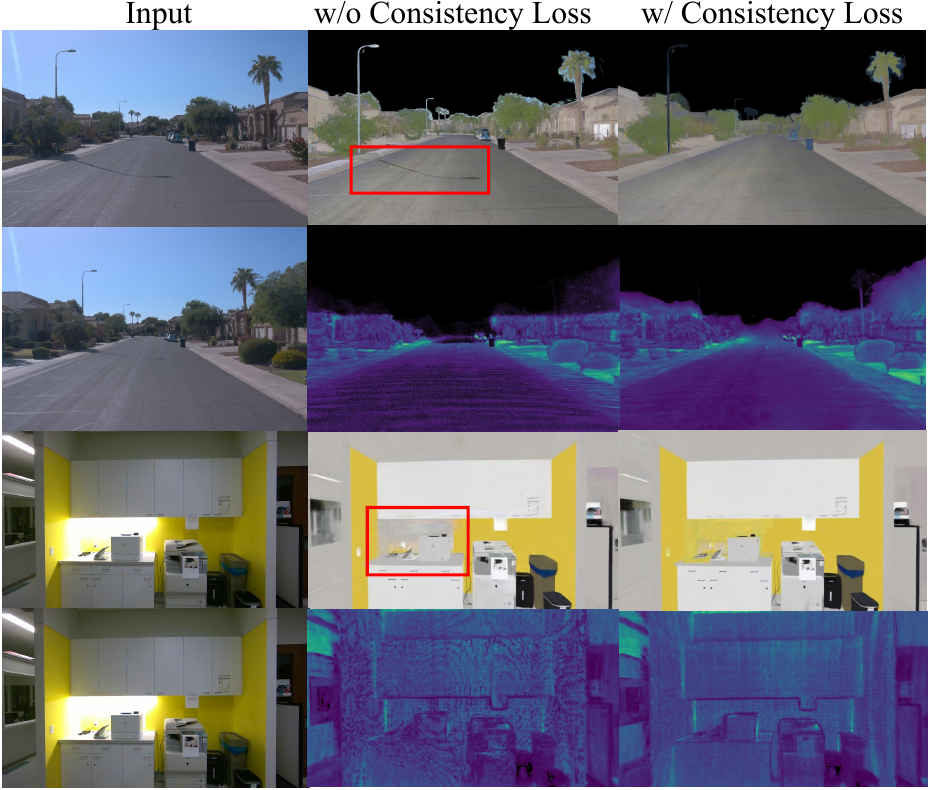}
  \caption{
    {\bf The RGB-LiDAR consistency loss corrects significant errors.}  Our proposed RGB-LiDAR consistency loss improves
    the robustness of surface reflectance estimation.  In each pair of rows, {\bf top} is spectral reflectance, {\bf
      bottom} is LiDAR albedo.  The cast shadow in the top pair is fixed, as is the color error around the laser printer in
    the second pair.
    }
  \label{fig: consitency_loss}
\end{figure}

\begin{figure*}[t]
  \centering
  \includegraphics[width=1\textwidth]{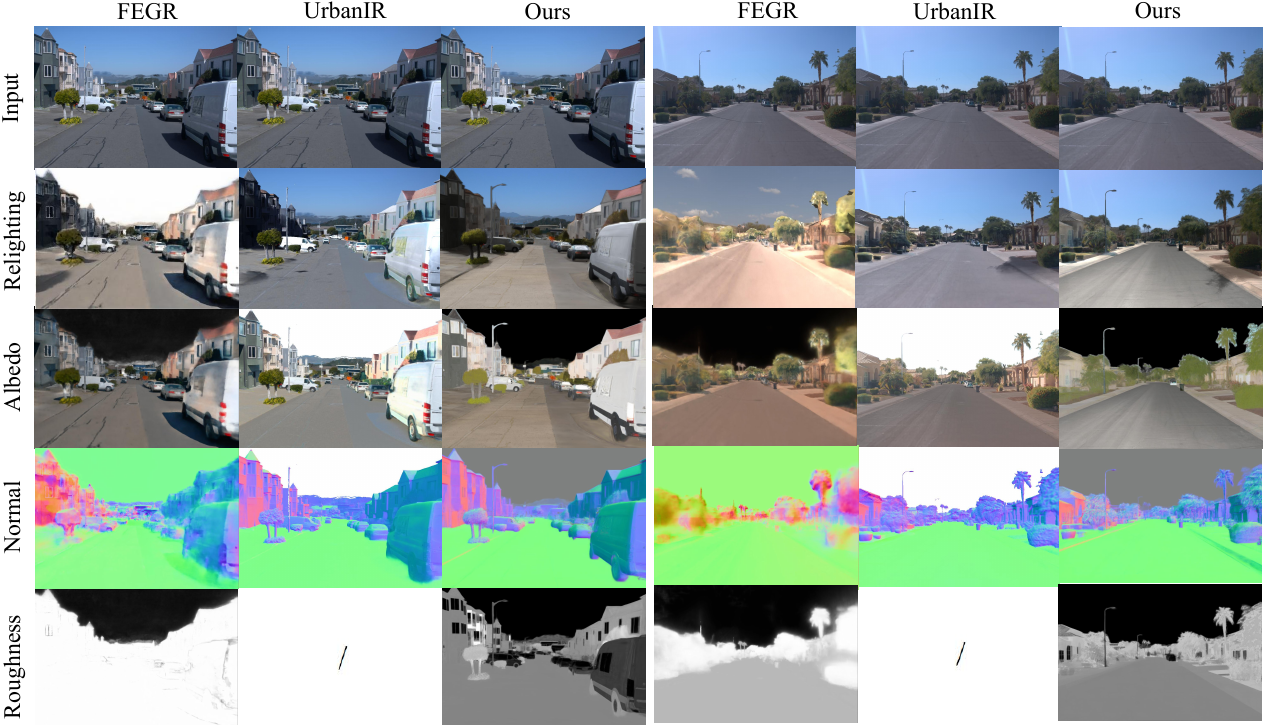}
  \caption{{\bf Qualitative comparison for inverse rendering with FEGR and UrbanIR on Waymo dataset.} FEGR produces
    unrealistic normal estimates and bakes hard shadows into the albedo. UrbanIR’s has no dense roughness estimation,
    and its radiance-based shadows cause relighting artifacts (see row 2, column 5). In contrast, our method achieves
    accurate and plausible material and geometry estimates, yielding superior relighting. Notice also the improved
    qualitative ``realism'' in relighting figures; surfaces tend to look more like actual object surfaces, and less like
    computer graphics items, likely a consequence of our roughness model.  Both FEGR and our
    method use LiDAR, while UrbanIR relies solely on video input. 
  }
  \label{fig:fegr_and_urbanir}
\end{figure*}

\subsection{Comparison with SOTA methods}

\paragraph{Inverse Rendering} We compare our method against UrbanIR \cite{lin2023urbanir} and FEGR \cite{wang2023neural}, two state-of-the-art approaches for urban scene inverse rendering. Since the Waymo dataset does not provide ground-truth intrinsic labels, we present only qualitative comparisons in Fig. \ref{fig:fegr_and_urbanir}, using baseline results provided by the authors of UrbanIR. Our method achieves superior inverse rendering by leveraging reflectance to effectively disentangle shading from albedo, resulting in smoother albedo estimates. In contrast, both UrbanIR and FEGR struggle to separate shadows cast by lighting poles and those beneath vehicles from the albedo, resulting in unrealistic shadows beside the car in the relighting results of the first scene.
Additionally, we compare our albedo estimation with RGB-X \cite{Zeng_2024} in Fig. \ref{fig:mono}, which serves as the diffusion-based prior for our framework. RGB-X suffers from multi-view inconsistency and fails to recover the albedo of cars. In contrast, our intrinsic 3D Gaussian representation is inherently multi-view and time-consistent, enabling it to correct erroneous predictions during training. 

\begin{table}[t]
    \centering
    \begin{tabular}{l|ccc}
        \toprule
        \textbf{Method} & PSNR↑ & SSIM↑ & LPIPS↓  \\
        \midrule
        UrbanIR  \cite{lin2023urbanir} & 28.84 & 0.67 & 0.49 \\
        \midrule
        w/o $\mathcal{L}_{lidar \rightarrow rgb}$ &  29.97 & \textbf{0.73} & 0.34 \\
        Ours  & \textbf{30.42} & 0.72  & \textbf{0.30}  \\
        \bottomrule
    \end{tabular}
    \caption{{\bf Quantitative results for relighting.} 
    }
    \label{tab: relight}
\end{table}

\begin{figure*}[t]
  \centering
  \includegraphics[width=1\linewidth]{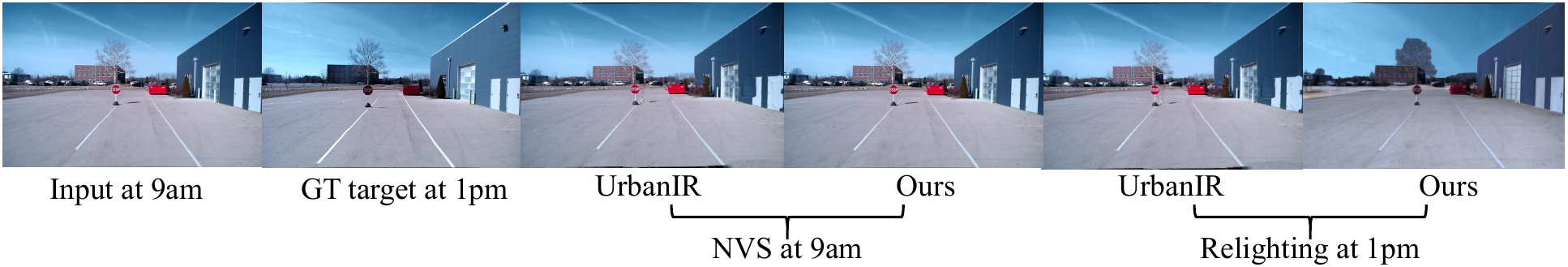}
  \caption{Qualitative results for relighting. By replacing the illumination of the 9 AM scene with that of the 1 PM, we can successfully shade the tree and buildings.
    }
  \label{fig: relight}
\end{figure*}

\paragraph{Relighting} For quantitative evaluation, we use data captured under different lighting conditions. Specifically, we record a scene at 9 AM and 1 PM on the same day, train both sequences independently using our framework, and then replace the illumination parameters of the 9 AM scene with those from the 1 PM scene. Table~\ref{tab: relight} presents the quantitative results, where our method significantly outperforms UrbanIR. Additionally, incorporating the consistency loss further enhances performance, primarily due to more accurate material estimation.
Fig.~\ref{fig: relight} shows the qualitative results, highlighting noticeable light and shadow shifts on road signs and distant buildings. In contrast, UrbanIR struggles to adjust the lighting. This demonstrates that our framework effectively disentangles illumination from albedo, enabling accurate modeling of shading variations under different lighting.



\begin{table}[t]
    \centering
    \begin{tabular}{l|cccc|c}
        \toprule
         &  \multicolumn{5}{c}{\textbf{ Intensity-RMSE $\downarrow$}}  \\
        \cmidrule{2-6} 
        & \multicolumn{4}{c|}{Scene ID} & \multirow{2}{*}{Average} \\
        \textbf{\centering Method} & 1 & 2 & 3 & 4 & \\
        \midrule
        LiDARsim  \cite{manivasagam2020lidarsim}  & 0.12  & 0.13 & \textcolor{blue}{\textbf{0.09}} & 0.14 &0.120  \\  
        PCGEN \cite{li2023pcgen} & 0.11 & 0.15 & \textcolor{blue}{\textbf{0.09}} & 0.15 & 0.125 \\
        AlignMiF \cite{tao2024alignmif} & \textbf{0.05} & \textcolor{blue}{\textbf{0.10}} & \textbf{0.05} & 0.09 & \textcolor{blue}{\textbf{0.073}} \\
        NFL \cite{huang2023neural} & \textcolor{blue}{\textbf{0.06}} & 0.13 & \textbf{0.05} & \textcolor{blue}{\textbf{0.08}}  & 0.080 \\
        \midrule
        Ours & \textcolor{blue}{\textbf{0.06}} & \textbf{0.08} & \textbf{0.05} &  \textbf{0.06} & \textbf{0.063} \\
        \bottomrule
    \end{tabular}
    \caption{{\bf Quantitative results for novel view synthesis of LiDAR intensity on the Waymo Open Dataset.} The highlighted metrics denote \textbf{Best} and \textcolor{blue}{\textbf{Second Best}}. The proposed method achieves the best results overall. 
    }
    \label{tab: lidar rmse}
\end{table}

\paragraph{LiDAR Simulation} To validate the effectiveness of our LiDAR intensity formulation and the accuracy of the generated intensity, we evaluate novel view synthesis for LiDAR intensity on the Waymo Dataset. We compare our approach against a series of LiDAR simulation works including LiDARSim \cite{manivasagam2020lidarsim}, PCGEN \cite{li2023pcgen}, AlignMiF \cite{tao2024alignmif} and NFL \cite{huang2023neural}. Following \cite{huang2023neural}, we conduct experiments on four scenes, using 50 frames from each sequence for training and selecting every 5th frame for validation. We use RMSE as the evaluation metric for intensity.  Table \ref{tab: lidar rmse} presents the quantitative results, where our method achieves the lowest RMSE, outperforming all baselines. This demonstrates that our formulation effectively captures the underlying physical phenomena, leading to more accurate LiDAR intensity modeling.

\subsection{Ablation Studies}

\paragraph{RGB-LiDAR Consistency Loss} Fig. \ref{fig: consitency_loss} shows qualitative comparison of inferred spectral
reflectance with and without consistency loss. {\bf Consistency removes shadows:}
The shadows of the light pole are incorrectly embedded into the spectral reflectance ({\bf first row}) when consistency
is not applied. The consistency loss recovers the road correctly.
{\bf Consistency improves LiDAR albedo:} The upper part of the images in the second row is missing LiDAR albedo
estimates when consistency is not applied, because the elevation range of the sensor is  limited.
The consistency loss propagates information from the spectral reflectance effectively propagated, filling in the missing bits.
{\bf Consistency fixes lighting induced errors:} The indoor scene of the third row shows a standard problem with
estimating RGB spectral reflectance from images: spatially fast changes in illuminant baffle all intrinsic image
methods, so some lighting effects get ``baked'' into results.  The consistency loss significantly improves the accuracy of the albedo estimation.  



\begin{figure*}[t]
  \centering
  \includegraphics[width=.8\textwidth]{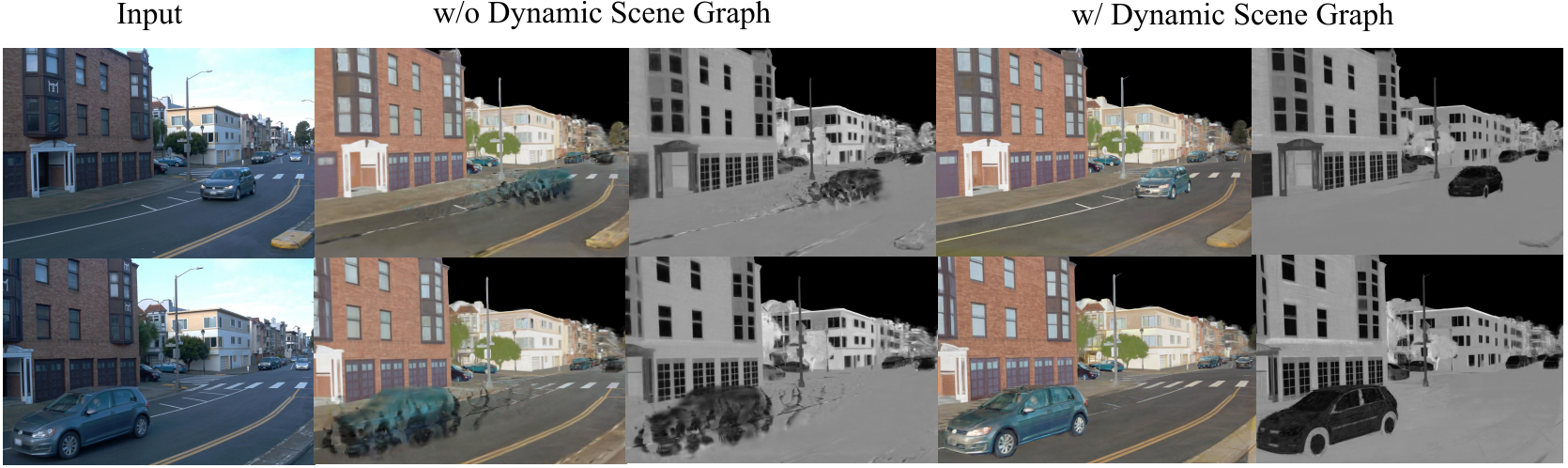}
  \caption{{\bf Ablation study on dynamic scene graph} Explicitly modeling dynamic objects improves albedo and roughness estimation; otherwise, motion-blurred artifacts will be baked into the scene.}
  \label{fig:dynamic}
\end{figure*}

\paragraph{Dynamic Scene Graph} We conduct an ablation study to assess the effectiveness of the dynamic scene graph by removing the dynamic nodes and training a static set of Gaussians on dynamic video input. Tab. \ref{tab: ablation} presents the quantitative results for the reconstruction of PBR images on a dynamic scene in the Waymo Dataset. Besides, Fig. \ref{fig:dynamic} presents the estimated albedo and roughness. As shown in the figure, without the dynamic scene graph, the moving car exhibits aliasing artifacts due to the inability to model motion. In contrast, our approach effectively captures time-varying intrinsic properties and handles the changing shadows beneath the car, enabling robust inverse rendering from dynamic video input.

\paragraph{LiDAR Input} To evaluate the impact of LiDAR data on our framework, we conduct an ablation study where only RGB images are used. Specifically, we disable the LiDAR-based initialization of 3D-GS and exclude the albedo consistency loss term from the optimization. The reconstruction performance is reported in Table~\ref{tab: ablation}.  While inverse rendering can still be performed without LiDAR sequence as inputs, the quality of physically based rendering (PBR) images exhibits a notable decrease. This study highlights the crucial role of LiDAR in enhancing both geometric fidelity and inverse rendering accuracy.

\begin{table}[t]
    \centering
    \begin{tabular}{l|ccc}
        \toprule
        \textbf{Method} & PSNR$\uparrow$ & SSIM$\uparrow$ & LPIPs$\downarrow$  \\
        \midrule
        w/o LiDAR     & 33.35  & 0.89 &  0.13\\
        w/o Dynamic & 29.13 & 0.83 & 0.21 \\
        \midrule
        Ours  & \textbf{34.76} & \textbf{0.91} & \textbf{0.11}   \\
        \bottomrule
    \end{tabular}
    \caption{{\bf Ablation studies on rendering quality.} The highlighted metrics denote \textbf{Best}. Both LiDAR reflectance and dynamic modeling improve reconstruction quality.}
    \label{tab: ablation}
\end{table}

\subsection{Downstream Applications}

\begin{figure}[t]
  \centering
  \includegraphics[width=1\linewidth]{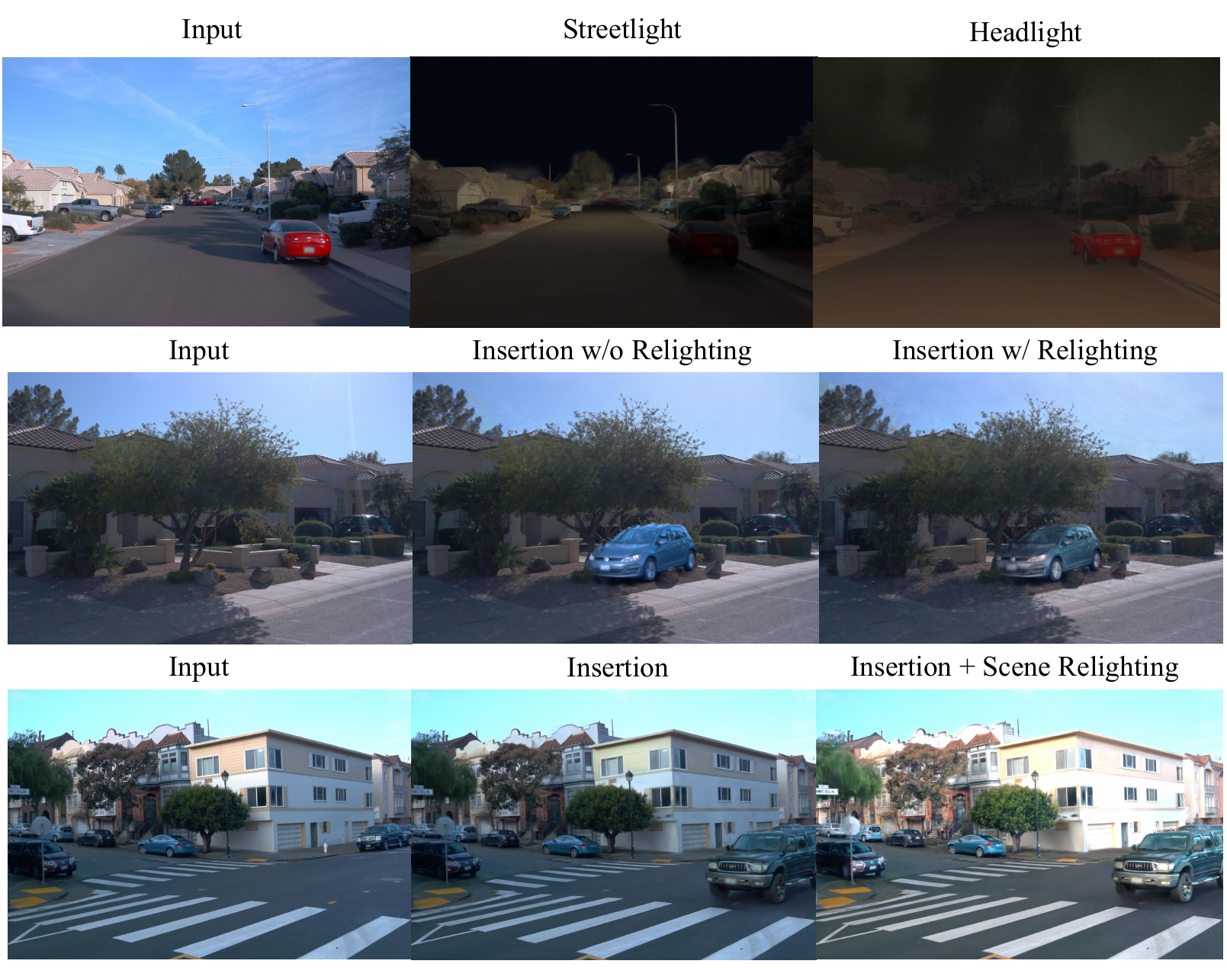}
  \caption{{\bf Downstream applications of our method}
Top: night simulation with controllable lights; middle: insertion with/without relighting; bottom: insertion rendering with/without changing the time of day.}
  \label{fig:editing}
\end{figure}

\paragraph{Scene Editing} Fig. \ref{fig:editing} presents the scene editing results, showing the versatility of our method in both relighting and object insertion. In the first row, we demonstrate nighttime simulation with streetlight and headlight illumination to an input daytime scene. Additionally, we present object insertion results. Unlike previous approaches \cite{lin2023urbanir}, which rely on off-the-shelf rendering engines like Blender \cite{blender}, we can directly transfer a trained dynamic node from one scene to another and relight the node using our framework. The second row shows the result without relighting the inserted node: the inserted car appears mismatched with the scene. In contrast, with relighting, the car blends seamlessly into the environment. The third row shows the results of relighting both the scene and the inserted object simultaneously.

\begin{table}[h]
    \centering
    \begin{tabular}{l|ccc}
        \toprule
        \textbf{Method} & Precision↑ & Recall↑ & mAP@50↑  \\
        \midrule
        w/o Night Aug.   & 0.537  &    0.281   &   0.236   \\
        w/ Night Aug. & \textbf{0.674}  &    \textbf{0.312}    &  \textbf{0.321}  \\
        \bottomrule
    \end{tabular}
    \caption{{\bf Data augmentation for nighttime object detection.} Off-the-shelf object detection~\cite{jocher2020ultralytics} does not perform well on Waymo nighttime sequences. Our night simulation generates nighttime logs from daytime labeled logs at no additional cost. 
    }
    \label{tab: night_sim}
\end{table}

\paragraph{Nighttime Data Augmentation for Object Detection} To evaluate the applicability of our method in autonomous driving perception, we conduct an experiment leveraging our method for nighttime data augmentation. Specifically, we transform daytime image sequences into nighttime conditions using our framework while preserving the original object detection labels. This enables the generation of nighttime training data without additional manual annotations. We generate 100 nighttime images with a total of 121 car labels which are then used to fine-tune a YOLO-v5 object detection model \cite{jocher2020ultralytics}. We evaluate the model using 50 real nighttime images from Waymo Dataset. Comparing its performance against the baseline without fine-tuning in Tab. \ref{tab: night_sim}, we demonstrate the potential to enhance nighttime perception for autonomous driving, particularly in low-visibility conditions, without the costly process of collecting and labeling nighttime data.

\section{Limitation and Discussion}

In this work, we integrate LiDAR into inverse rendering and introduce \ours, novel model capable of reconstructing large-scale, relightable, and dynamic scenes from a single RGB+LiDAR sequence. By leveraging the consistency between LiDAR and RGB albedo, our approach enhances material estimation and enables a variety of scene editing applications, including relighting, object insertion, and nighttime simulation. However, there are still limitations. First, we adopt a BVH-based ray tracer for 3D Gaussian ray tracing, which can produce inaccurate shadows due to the opacity properties of Gaussian primitives. Additionally, our illumination model, which accounts for only skylight and sunlight, is not sufficient for inverse rendering on complex environments such as nighttime scenes, which we will try to address in the future.

%% file: sec/6_supplementary.tex
\appendix

\section{Appendix}

\subsection{Custom Data Collection}
\begin{figure}[h!]
\begin{center}
\includegraphics[width=0.47\linewidth]{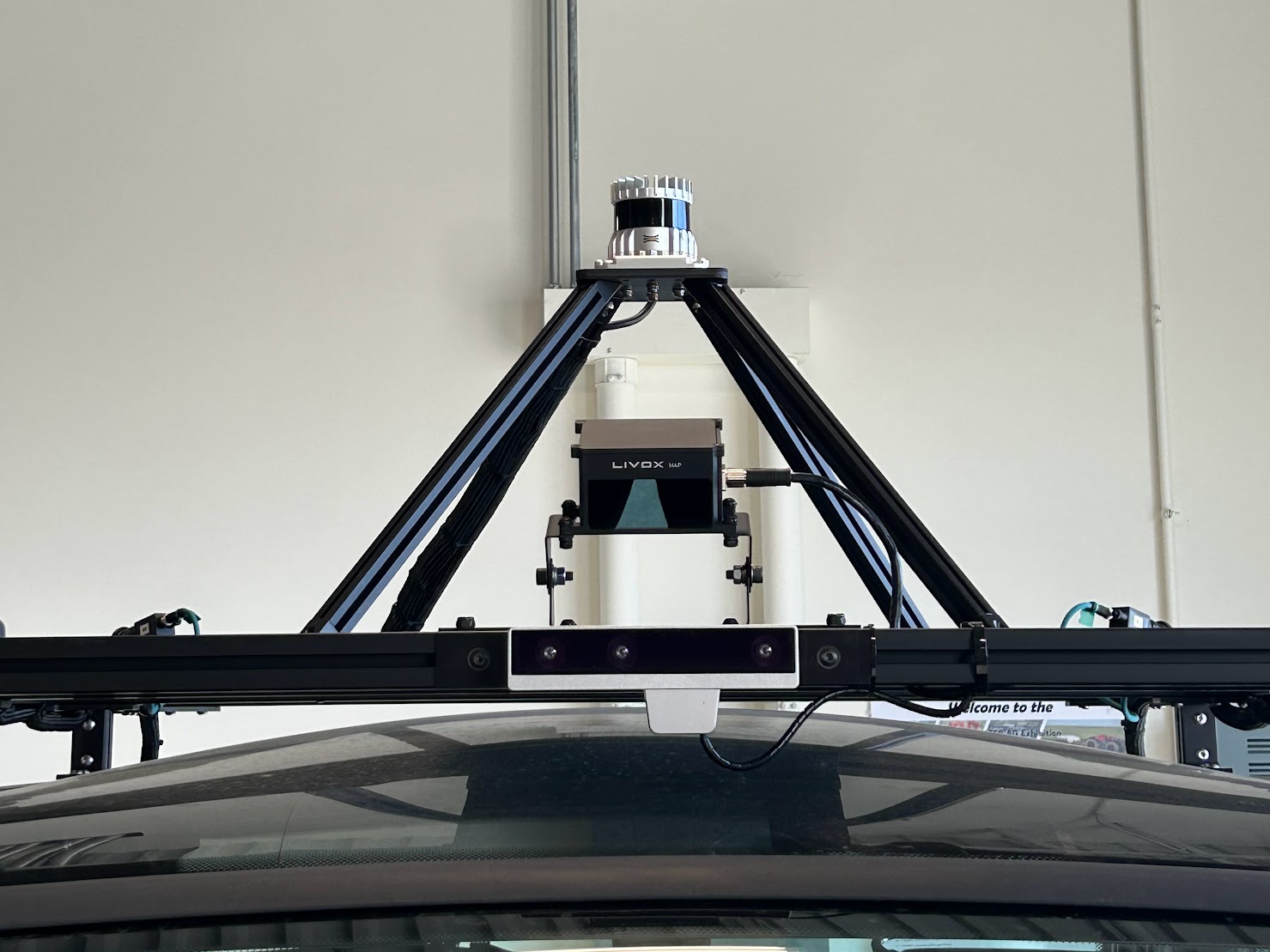} \hspace{0.1em}
\includegraphics[width=0.47\linewidth]{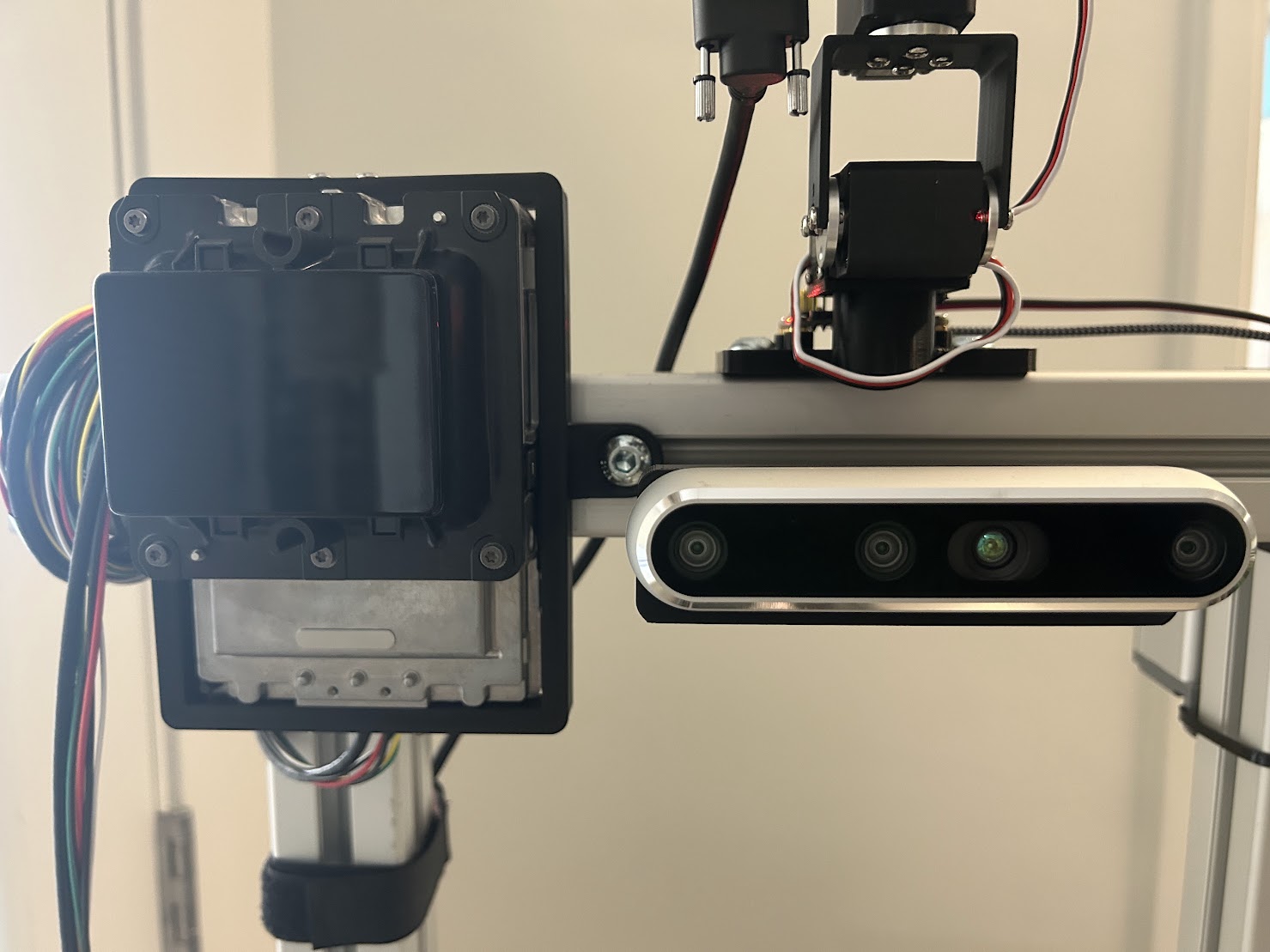}
\end{center}
\vspace{-5mm}
\caption{The two LiDAR-camera systems used for data collection. The left is for outdoor, and the right is for indoor.}
\label{fig:lidar_camera_systems}
\end{figure}

We recorded data ourselves using two LiDAR-camera systems(shown in Fig.~\ref{fig:lidar_camera_systems}) to enable our outdoor quantitative relighting evaluation and indoor albedo-reflectance experiments. In both cases, we used targetless LiDAR-camera calibration~\cite{koide2023lidarcameracalib} to obtain the coordinate transform between the LiDAR and camera.

\paragraph{Outdoor} We recorded data from the Polaris Gem e4, a street-legal, four-seater vehicle outfitted with RTK GPS, an Ouster OS1-128 LiDAR, and an Oak-D LR stereo camera. The experiments were carried out in a shared testing track facility with a secure testbed area. A designated safety driver
and safety lookout were present at all times. We collected the dataset by manually driving the vehicle along the same trajectory/scene at different times throughout the day.

\paragraph{Indoor} We used an AgileX Ranger Mini 2 mobile robot platform with a Hesai FT120 solid-state LiDAR and Realsense D455 depth camera mounted on top. We recorded experiments by teleoperating the robot inside an academic building.

\subsection{LiDAR Intensity Reveals Specularity}
\begin{figure}[t]
\begin{center}
\includegraphics[width=0.95\linewidth]{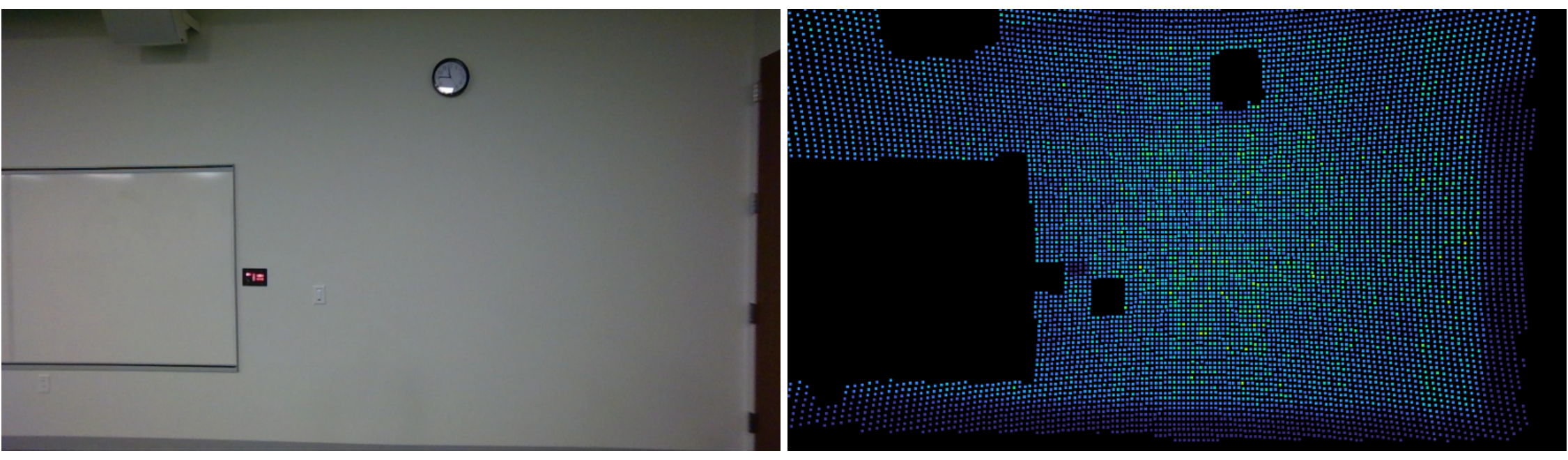}
\\[-1em]
\includegraphics[width=0.95\linewidth]{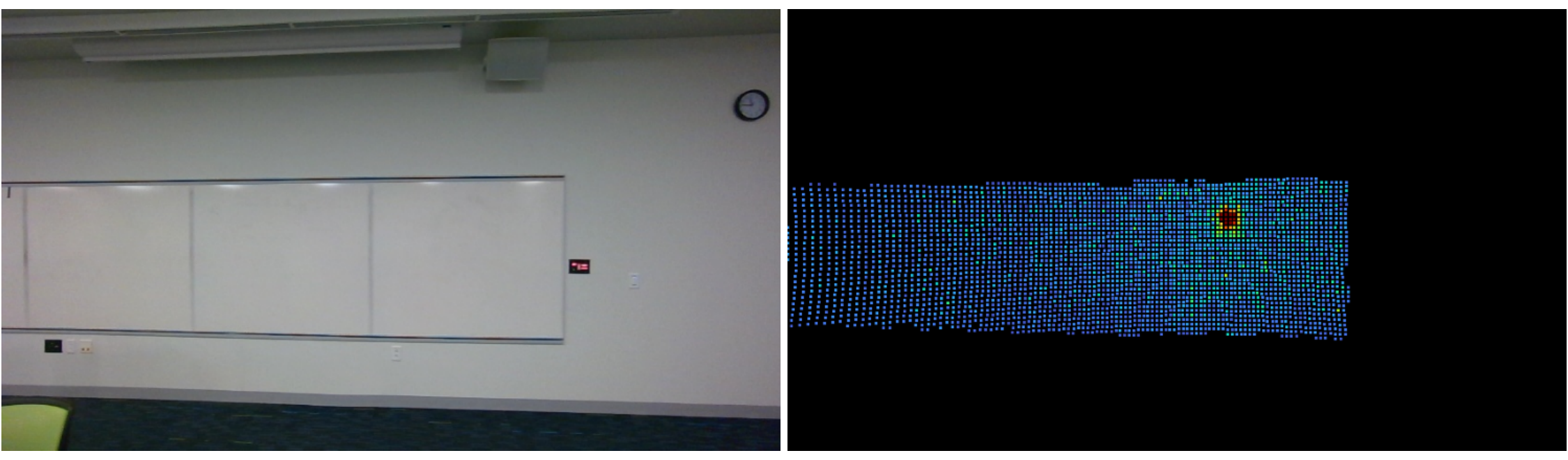}
\end{center}
\vspace{-5mm}
\caption{RGB and masked lidar intensity for a diffuse wall(top) and a specular whiteboard(bottom).}
\label{fig:lidar_specularity}
\end{figure}

We verify that LiDAR intensity can indicate surface specularity by collecting real-world data of a diffuse wall and a specular whiteboard. In Fig.~\ref{fig:lidar_specularity}, the specular whiteboard has a high intensity only around the center of the image where the LiDAR beams are parallel to the surface normal of the whiteboard. Since it is a specular surface, those parallel LiDAR beams are reflected back at the same angle, straight into the sensor, and the LiDAR receiver gets the strongest signal/highest intensity in that region. The diffuse wall, however, shows no major intensity difference and is roughly uniform across all the LiDAR points since it reflects light in all directions. 

\begin{figure}[t]
\begin{center}
\includegraphics[width=1.0\linewidth]{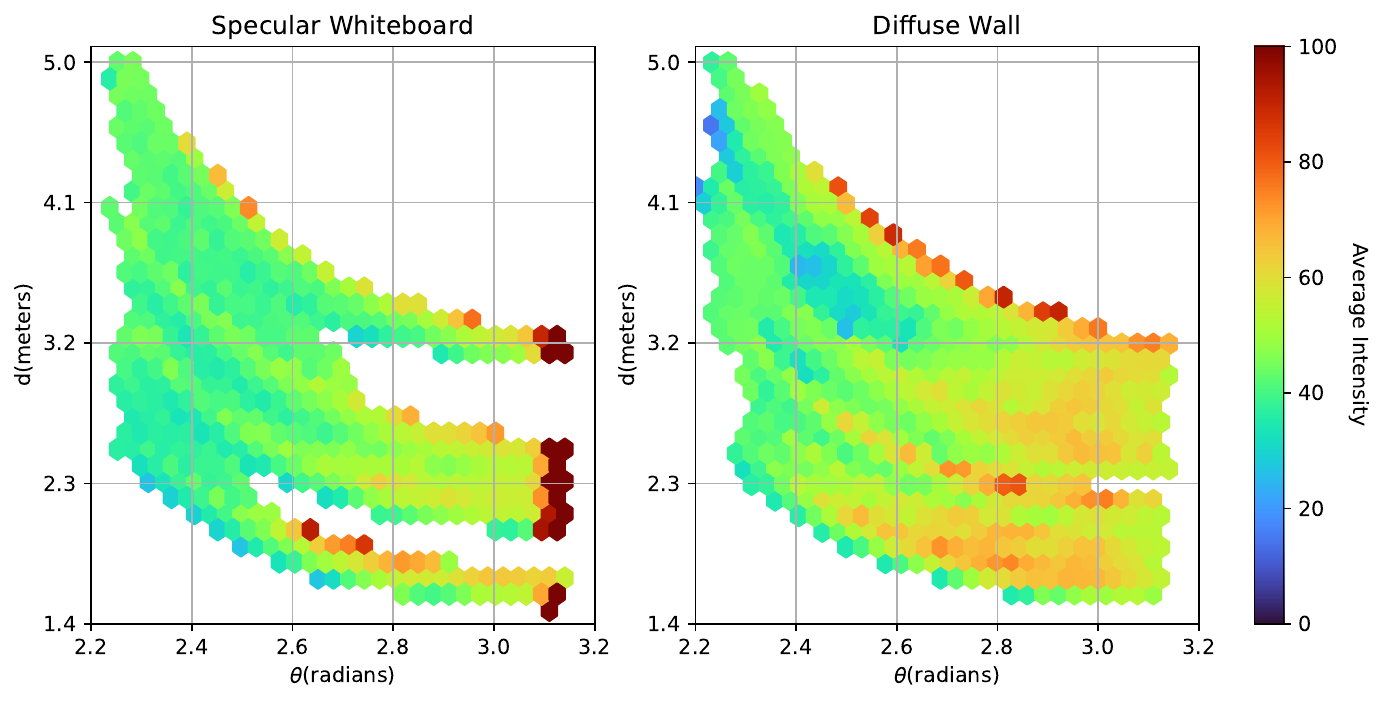}
\end{center}
\vspace{-5mm}
\caption{LiDAR intensity is visualized for two different objects: a specular whiteboard and a diffuse wall. The x-axis, $\theta$, is the angle between a LiDAR ray and the corresponding surface normal. The y-axis, d, is the distance to the LiDAR point. Each hexbin represents the average LiDAR intensity among all the LiDAR points within that bin.}
\label{fig:lidar_intensity_plot}
\end{figure}

In Fig.~\ref{fig:lidar_intensity_plot}, we show this continues to hold true at varying distances and reflectance angles. We recorded a sequence of data scanning both objects and accumulated the intensities for points corresponding to each one. For the specular whiteboard, the highest intensity LiDAR points are clustered around $\theta=3.14$, when the LiDAR ray and surface normal are parallel. But for the diffuse wall, the intensity is roughly spread out as expected. This shows how LiDAR intensity can be a valuable cue to determine specularity.

\subsection{Physics-based LiDAR Reflectance Model}

Here, we provide the mathematical derivation for the specular term of the LiDAR Reflectance Model. The Cook-Torrance BRDF model \cite{cook1982reflectance} for the specular component is given by:

\begin{align}
f_s(\mathbf{\omega_i}, \mathbf{\omega_o}) &= \frac{F(\mathbf{\omega_i}) \, G(\mathbf{\omega_i}, \mathbf{\omega_o}) \, D(\mathbf{h})}{4 (\mathbf{\omega_i} \cdot \mathbf{n}) (\mathbf{\omega_o} \cdot \mathbf{n})} \notag \\ 
&= \frac{F(\mathbf{\omega_i}) \, G(\mathbf{\omega_i}, \mathbf{\omega_o}) \, D(\mathbf{h})}{4 \cos^2\theta}, \notag
\end{align}

where:
\begin{itemize}
    \item \( D(\mathbf{h}) \) is the microfacet distribution function, modeling the distribution of surface normals, with \( \mathbf{h} \) being the half-angle vector.
    \item \( G(\mathbf{\omega_i}, \mathbf{\omega_o}) \) is the geometry term, accounting for masking and shadowing of microfacets.
    \item \( F(\mathbf{\omega_i}) \) is the Fresnel term, which models reflectance based on the viewing angle.
\end{itemize}

For the LiDAR lighting model, where \( \mathbf{\omega_i} = \mathbf{\omega_o} \), the half-angle vector simplifies to:
\[
\mathbf{h} = \frac{\mathbf{\omega_i} + \mathbf{\omega_o}}{\|\mathbf{\omega_i} + \mathbf{\omega_o}\|} = \omega_o.
\]

The microfacet distribution function is commonly modeled using the GGX distribution:

\begin{align}
D(\mathbf{h}) &= \frac{\alpha^2}{\pi \left((\mathbf{h} \cdot \mathbf{n})^2 (\alpha^2 - 1) + 1\right)^2} \notag \\
&= \frac{\alpha^2}{\pi \left( \cos^2\theta (\alpha^2 - 1) + 1\right)^2}, \notag
\end{align}

where \( \alpha \) represents the surface roughness.

For the Fresnel term, we use:

\[
F(\mathbf{\omega_i}) = F_0 + (1 - F_0)(1 - (\mathbf{\omega_i} \cdot \mathbf{h}))^5 = F_0.
\]

For the geometry term, following the Cook-Torrance BRDF model, it can be calculated as:

\begin{align}
G &= \min\left(1, \frac{2 (\mathbf{\omega_i} \cdot \mathbf{n}) (\mathbf{n} \cdot \mathbf{h})}{(\mathbf{\omega_o} \cdot \mathbf{h})}, \frac{2 (\mathbf{\omega_o} \cdot \mathbf{n}) (\mathbf{n} \cdot \mathbf{h})}{(\mathbf{\omega_o} \cdot \mathbf{h})}\right) \notag \\
&= \min(1, 2\cos^2\theta). \notag
\end{align}

Thus, the final specular term is:

\begin{align}
f_s(\mathbf{\omega_i}, \mathbf{\omega_o}) &= \frac{F(\mathbf{\omega_i}) \, G(\mathbf{\omega_i}, \mathbf{\omega_o}) \, D(\mathbf{h})}{4 \cos^2\theta} \notag \\
&= \frac{F_0 \alpha^2 \min(1, 2\cos^2\theta)}{4 \pi \cos^2\theta \left( \cos^2\theta (\alpha^2 - 1) + 1\right)^2}. \notag
\end{align}

\begin{figure*}[t]
\begin{center}
\includegraphics[width=.9\linewidth]{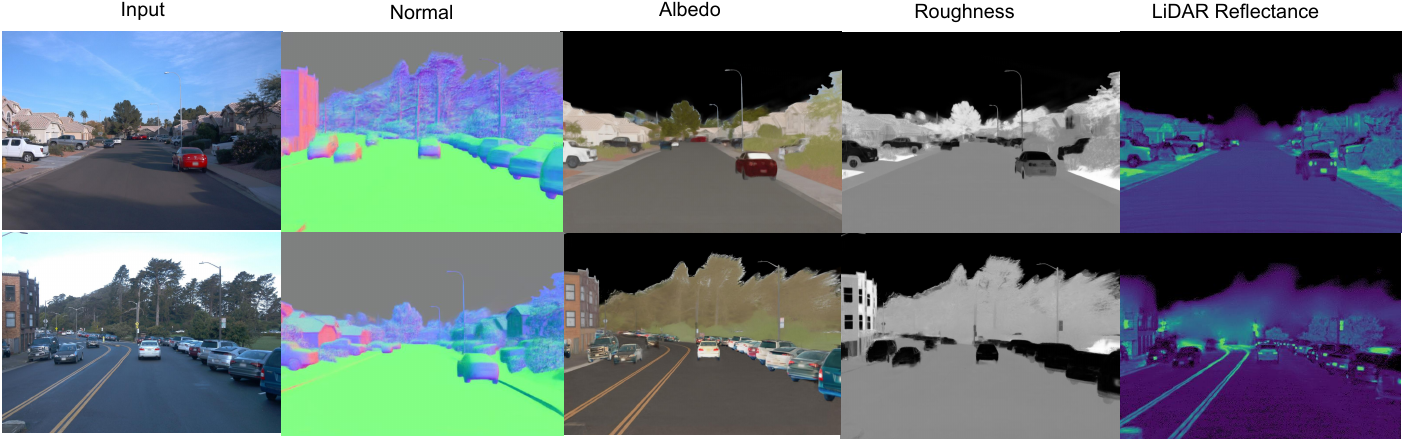}
\end{center}
\vspace{-5mm}
\caption{More results for inverse rendering.}
\label{fig:more1}
\end{figure*}

\begin{figure}
\begin{center}
\includegraphics[width=1.0\linewidth]{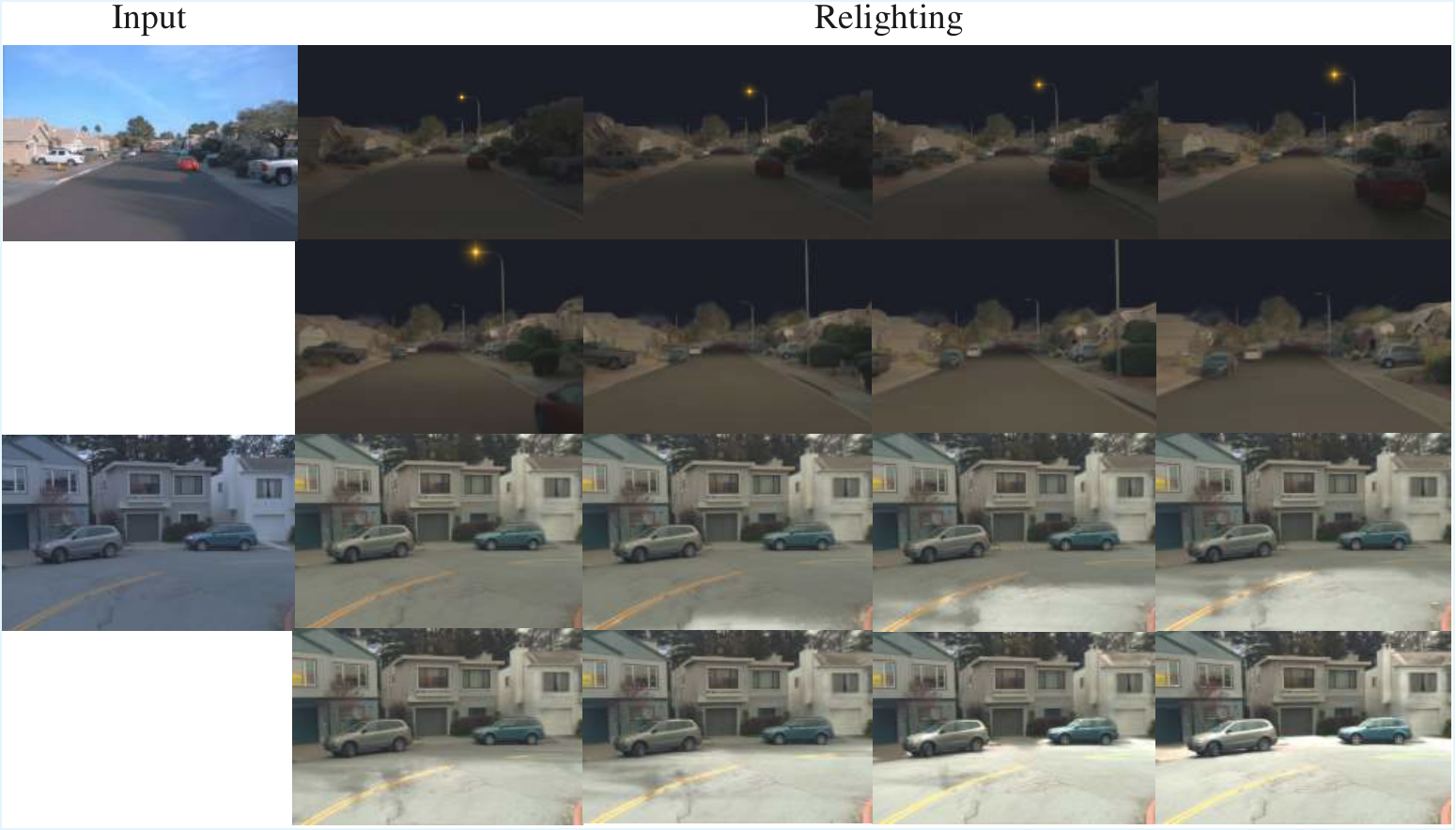}
\end{center}
\vspace{-5mm}
\caption{Relighting results on video sequences.}
\label{fig:relight}
\end{figure}

\begin{figure}
\begin{center}
\includegraphics[width=1.0\linewidth]{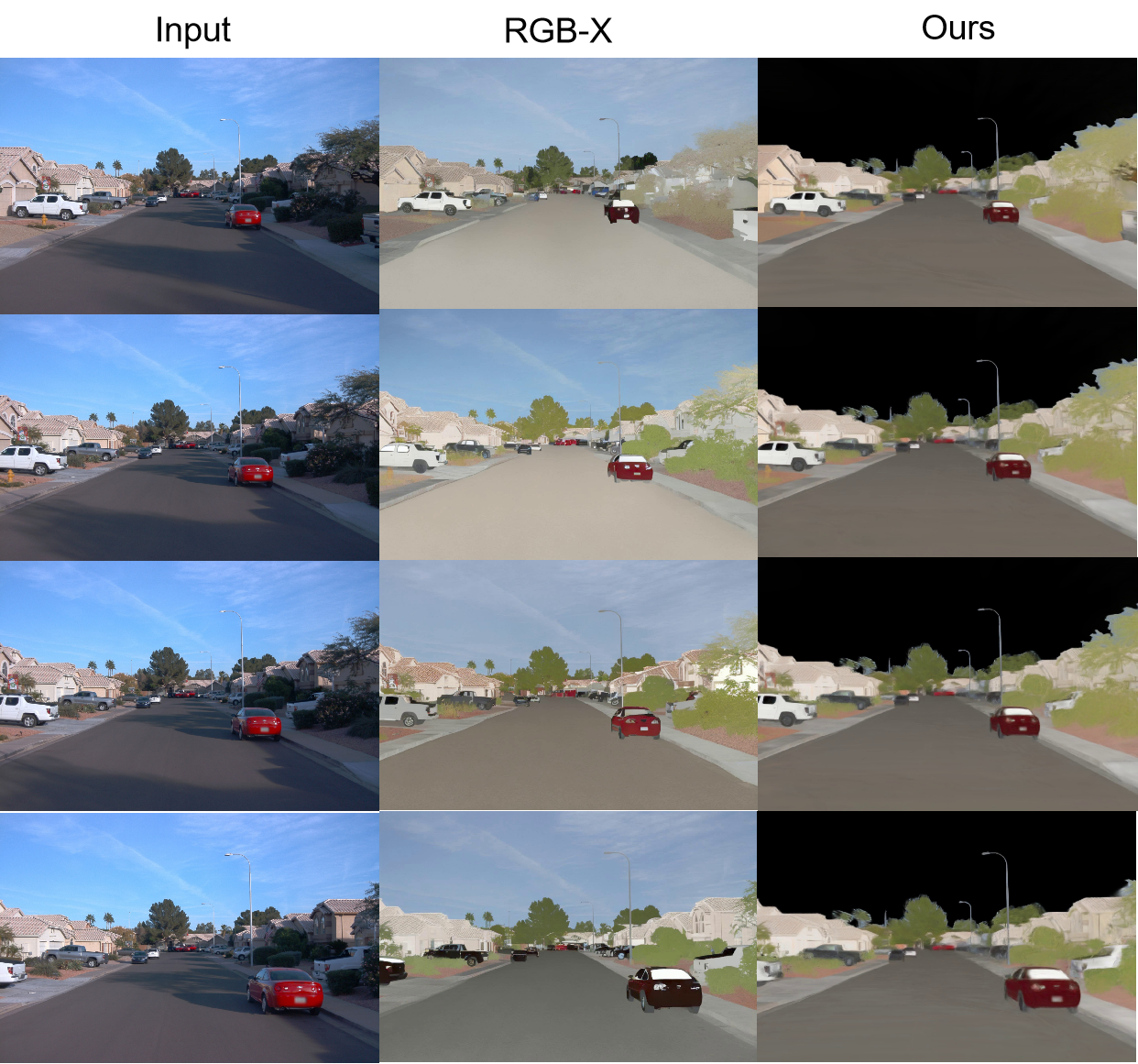}
\end{center}
\vspace{-5mm}
\caption{Comparison for albedo estimation with RGB$\leftrightarrow$X.}
\label{fig:more2}
\end{figure}

\subsection{Implementation Details}

\paragraph{Data Preprocessing} For image inputs, we preprocess each frame and acquire diffusion-based priors for materials and normals using RGB$\leftrightarrow$X \cite{Zeng_2024} and GeoWizard \cite{fu2024geowizard}. For the indoor scene, we obtain an additional lighting mask $\mathbf{M}_\mathrm{light}$ by filtering the luminance. This mask is used to exclude unreliable regions from the rendering loss, where high-intensity lighting makes it unreliable to estimate albedo.

For LiDAR sequences, we project the LiDAR points into image space using the camera-LiDAR pose transformation, resulting in a sparse LiDAR intensity map. The intensity values are then normalized to the range [0, 1]. Furthermore, the Waymo dataset provides the ground truth object poses for each object, which are used as $\mathbf{T}_k(t)$ to transform dynamic nodes into the world coordinate system.

\paragraph{Method Details} We develop our framework based on OmniRe \cite{chen2024omnire}, a 3D-GS framework designed for driving scene. We initialize the means of the background 3D-GS with LiDAR points. Specifically, we set the maximum point number to $8 \times 10^5$. If the number of LiDAR points exceeds this limit, we randomly sample points. For rigid nodes, we randomly sample 5,000 points for initialization within the 3D bounding boxes.

For the camera rendering process, we adopt Monte Carlo ray tracing. For each Gaussian primitive $g$, we generate $M$ incident ray directions using Fibonacci sphere sampling based on the normal direction. During training, we set $M = 16$, while for inference, we use $M = 128$.

\paragraph{Optimization}  The model is trained for 30,000 iterations with a single NVIDIA A100 GPU. It takes approximately 2-3 hours of training for each scene. The learning rate is set to $1 \times 10^{-5}$. We adopt a two-stage training procedure: in the first stage, which consists of the first 15,000 iterations, we follow the general 3D-GS split replication approach, and all intrinsic properties are optimized simultaneously. After completing the first stage, the LiDAR albedo is processed into masks, and we stop duplicating and deleting 3D-GS nodes. In the next 15,000 iterations, all intrinsic properties except for LiDAR albedo and RGB albedo are fixed, and lighting conditions are also optimized. The total loss is defined as:
\begin{align}
\mathcal{L}_{total} = \lambda_1 \mathcal{L}_{lidar} + \lambda_2 \mathcal{L}_{rgb} + \lambda_3 \mathcal{L}_{nor} +  \lambda_4 \mathcal{L}_{mat} \notag \\
+  \lambda_5 \mathcal{L}_{rgb \rightarrow lidar}  +  \lambda_6 \mathcal{L}_{lidar \rightarrow rgb},
\end{align}

where $\lambda_i$ is the loss weight for each term. Specifically, we set $\lambda_1 = \lambda_2 = 1$, $\lambda_3 = \lambda_4 = 0.1$, and $\lambda_5 = \lambda_6 = 0.05$.

\paragraph{Application} For object insertion, we transfer the trained dynamic nodes from the Waymo dataset into new scenes by directly loading the corresponding checkpoints. Since the dynamic nodes retain their intrinsic properties, a relit result can be directly obtained through our camera rendering process.  For nighttime simulation, we remove the sunlight representation and set the sky dome lighting to a small constant intensity. Additionally, we use a spotlight to model the headlights and streetlights, with its center positioned at the camera origin or the light pole. The light intensity decreases with the square of the distance from the spotlight center to the Gaussian's means.

\begin{figure}[t]
  \centering
  \includegraphics[width=1\linewidth]{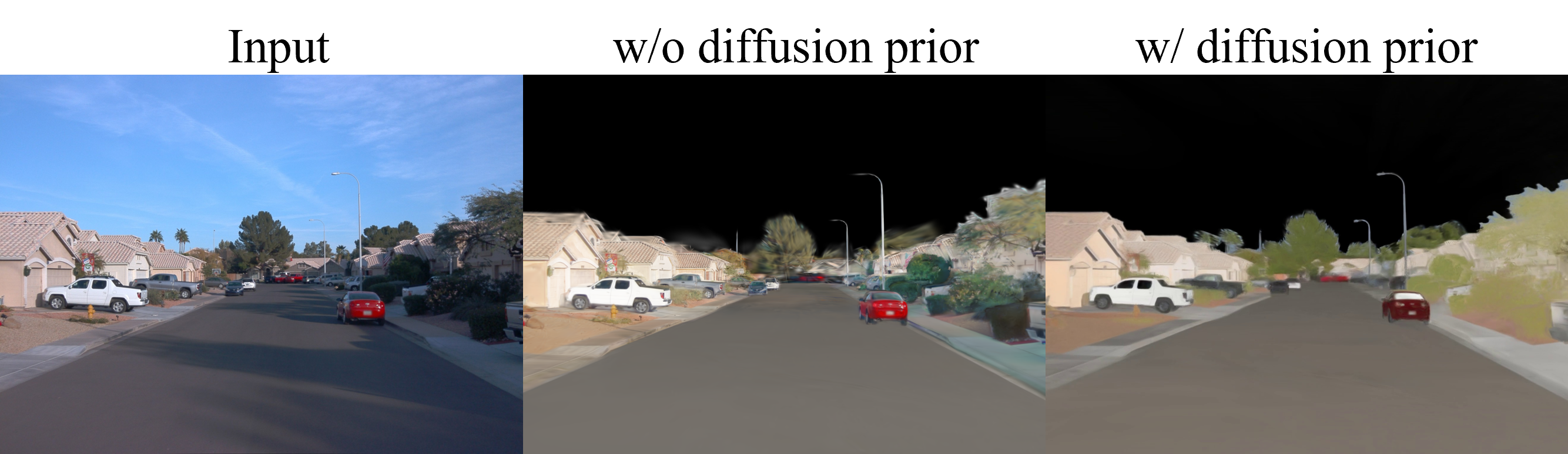}
  \vspace{-3mm}
  \caption{Ablation study on diffusion prior.}
  \label{fig:diffusion}
\vspace{-3mm}
\end{figure}

\subsection{More Qualitative Results}

\paragraph{More Inverse Rendering} Fig.~\ref{fig:more1} presents additional inverse rendering results, including normal, albedo, roughness, and LiDAR reflectance. As shown in the figure, the LiDAR and RGB albedo are consistent, and we can successfully disentangle shadows from the albedo.

\paragraph{More Relighting Results} We perform relighting on two sequences: the first simulates nighttime conditions, and the second continuously changes the sun direction. We present 8 frames of each video in Fig.~\ref{fig:relight}.

\paragraph{Comparison with Diffusion Prior} Fig.~\ref{fig:more2} compares the albedo estimation results of our method against RGB$\leftrightarrow$X on an image sequence. The albedo prior from RGB$\leftrightarrow$X exhibits significant temporal inconsistency due to the inherent limitations of the monocular diffusion model. In contrast, our framework achieves time-consistent albedo estimation, highlighting the advantages of physically based optimization.

\paragraph{Ablation on Diffusion Prior} We conduct an ablation study to assess the role of the diffusion prior. As shown in Fig.~\ref{fig:diffusion}, incorporating the diffusion prior produces smoother albedo estimates and reduces shadow ambiguity (e.g., on the trees), highlighting its effectiveness.

\begin{figure}[t]
  \centering
  \includegraphics[width=1\linewidth]{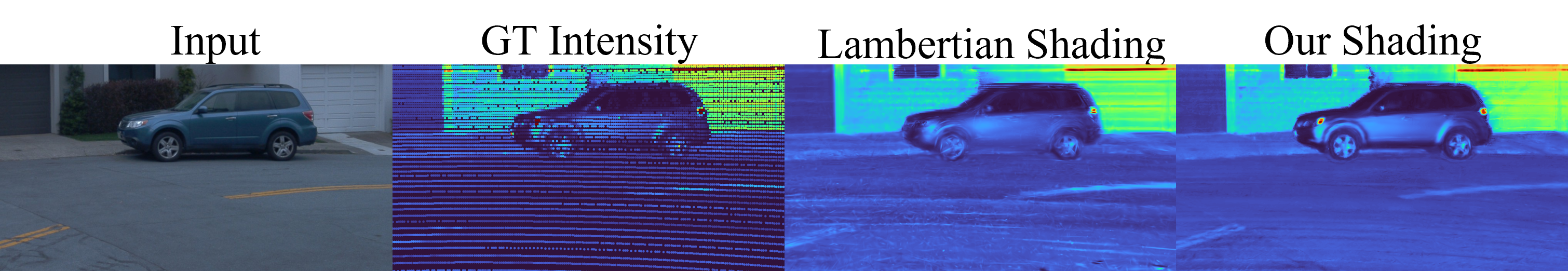}
  \vspace{-3mm}
  \caption{Ablation study on LiDAR reflectance modeling.}
  \label{fig:lidar_}
\vspace{-3mm}
\end{figure}

\paragraph{Ablation on LiDAR Reflectance Modeling} Tab.~\ref{tab: pbr lidar} and Fig.~\ref{fig:lidar_} provide  ablation studies of LiDAR simulation on a scene. Our reflectance model faithfully captures the specular highlights in the GT intensity—e.g., around the car’s front light and wheel arch—whereas the Lambertian model produces overly diffuse, physically implausible shading.

\begin{table}[t]
    \centering
    \begin{tabular}{l|ccc}
        \toprule
        \textbf{Method} & Intensity-RMSE $\downarrow$  \\
        \midrule\
        Lambertian  &  0.0493 \\
        Ours (PBR) & \textbf{0.0470}  \\
        \bottomrule
    \end{tabular}
    \vspace{-3mm}
    \caption{Ablation study on LiDAR reflectance modeling.}
    \label{tab: pbr lidar}
\end{table}